\PassOptionsToPackage{dvipsnames,table,xcdraw}{xcolor}
\documentclass[pdflatex,sn-mathphys-num]{sn-jnl}%
\setcitestyle{super,open={},close={}} %
\usepackage[arxiv]{optional}
\usepackage{silence}
\usepackage{enumerate}
\usepackage{latexsym}
\usepackage{graphicx}
\usepackage{multicol,multirow}
\usepackage{amsmath,amssymb,amsfonts}
\usepackage{amsthm}
\usepackage{appendix}

\usepackage[T1]{fontenc}
\usepackage{times}
\usepackage{textcomp}%
\usepackage{url}
\usepackage{comment}
\usepackage{xspace}
\usepackage{mathtools}
\usepackage{afterpage}
\usepackage{capt-of}
\usepackage{caption}
\usepackage[capitalize]{cleveref}

\crefname{figure}{Figure}{Figures}
\crefname{equation}{}{}
\usepackage{autonum}
\usepackage{booktabs}
\usepackage{gensymb}
\usepackage{algorithm}
\usepackage[noend]{algorithmic} %
\usepackage{soul}
\usepackage{tabularx}
\usepackage[export]{adjustbox}
\usepackage{enumitem}

\usepackage{multibib}
\newcites{methods}{Methods References}
\newcites{extended}{Extended Data References}

\usepackage{tikz}
\usetikzlibrary{calc}

\usepackage{lineno} %

\usepackage[normalem]{ulem} %

\graphicspath{{nature_figures/}{figures/}{figures/model_schematics}{figures/ecmwf}{figures/aifs}{figures/poet}{figures/fuxi}{figures/bias}{figures/extremes}{figures/aiwq}}

\DeclarePairedDelimiter\floor{\lfloor}{\rfloor}
\def\balign#1\ealign{\begin{align}#1\end{align}}
\def\baligns#1\ealigns{\begin{align*}#1\end{align*}}
\def\balignat#1\ealign{\begin{alignat}#1\end{alignat}}
\def\balignats#1\ealigns{\begin{alignat*}#1\end{alignat*}}
\def\bitemize#1\eitemize{\begin{itemize}#1\end{itemize}}
\def\benumerate#1\eenumerate{\begin{enumerate}#1\end{enumerate}}

\newenvironment{talign*}
 {\let\displaystyle\textstyle\csname align*\endcsname}
 {\endalign}
\newenvironment{talign}
 {\let\displaystyle\textstyle\csname align\endcsname}
 {\endalign}

\def\balignst#1\ealignst{\begin{talign*}#1\end{talign*}}
\def\balignt#1\ealignt{\begin{talign}#1\end{talign}}
\newcommand{\qtext}[1]{\quad\text{#1}\quad} 
 
\newcommand{\sstext}[1]{\ \ \text{#1}\ \ } 

\let\originalleft\left
\let\originalright\right
\renewcommand{\left}{\mathopen{}\mathclose\bgroup\originalleft}
\renewcommand{\right}{\aftergroup\egroup\originalright}

\def\tinycitep*#1{{\tiny\citep*{#1}}}
\def\tinycitealt*#1{{\tiny\citealt*{#1}}}
\def\tinycite*#1{{\tiny\cite*{#1}}}
\def\smallcitep*#1{{\scriptsize\citep*{#1}}}
\def\smallcitealt*#1{{\scriptsize\citealt*{#1}}}
\def\smallcite*#1{{\scriptsize\cite*{#1}}}

\def\mbi#1{\boldsymbol{#1}} %
\def\mbf#1{\mathbf{#1}}
\def\mbb#1{\mathbb{#1}}
\def\mc#1{\mathcal{#1}}

\def\reals{\mathbb{R}} %

\def\<{\left\langle} %
\def\>{\right\rangle}

\def\indic#1{\mbb{I}\left[{#1}\right]} %
\providecommand{\argmin}{\mathop\mathrm{arg min}}

\ifdefined\nonewproofenvironments\else
\ifdefined\ispres\else

\newenvironment{proof-sketch}{\noindent\textbf{Proof Sketch}
  \hspace*{1em}}{\qed\bigskip\\}
\newenvironment{proof-idea}{\noindent\textbf{Proof Idea}
  \hspace*{1em}}{\qed\bigskip\\}
\newenvironment{proof-of-lemma}[1][{}]{\noindent\textbf{Proof of Lemma {#1}}
  \hspace*{1em}}{\qed\\}
\newenvironment{proof-of-theorem}[1][{}]{\noindent\textbf{Proof of Theorem {#1}}
  \hspace*{1em}}{\qed\\}
\newenvironment{proof-attempt}{\noindent\textbf{Proof Attempt}
  \hspace*{1em}}{\qed\bigskip\\}

\fi

\fi
\newcommand{\pp}{\texttt{++}}
\newcommand{\perpp}{Persistence\pp\xspace}
\newcommand{\debpp}{Debias\pp\xspace}

\newcommand{\algorithmicinput}{\textbf{input}}
\newcommand{\INPUT}{\item[\algorithmicinput]}
\newcommand{\algorithmicoutput}{\textbf{output}}
\newcommand{\OUTPUT}{\item[\algorithmicoutput]}

\newcommand{\INIT}{\ENSURE}

\newcommand{\fcst}{\mbf{F}}
\newcommand{\leads}{\mathcal{L}}
\newcommand{\lstar}{\ell^\star}
\newcommand{\climvec}{\mbf{C}} 

\newcommand{\dstar}{d^{\star}}
\newcommand{\tstar}{t^{\star}}

\newcommand{\ind}{\mathbf{O}}
\newcommand{\gt}{\mathbf{y}}

\newcommand{\monthday}{\operatorname{monthday}} %
\newcommand{\myyear}{\operatorname{year}} 
\newcommand{\trainset}{\mc{T}}
\newcommand{\testset}{\mc{T}^\star}
\newcommand{\clip}{\operatorname{clip}}
\newcommand{\bss}{\textup{BSS}}
\newcommand{\bscore}{\textup{BS}}
\newcommand{\bsc}{\overline{\textup{BS}}}
\newcommand{\srpss}{\textup{Spatial-RPSS}}

\newcommand{\rpss}{\textup{RPSS}}
\newcommand{\rps}{\textup{RPS}}
\newcommand{\rpsc}{\overline{\textup{RPS}}}
\newcommand{\latitude}{\operatorname{latitude}}
\newcommand{\qobs}{q}
\newcommand{\qmodel}{q^{\operatorname{model}}}
\newcommand{\qecmwf}{q^{\operatorname{ecmwf}}}
\newcommand{\qfuxi}{q^{\operatorname{fuxi}}}
\newcommand{\qaifs}{q^{\operatorname{aifs}}}

\newcommand{\basesignificance}[1][all baselines]{rosshatching indicates significant improvement ($p < 0.05$) over {#1}.\xspace}
\newcommand{\significance}[1][all baselines]{C\basesignificance[#1]}
\newcommand{\lowercasesignificance}[1][all baselines]{c\basesignificance[#1]}

\newcommand{\singlesignificance}[2][one baseline]{Crosshatching indicates significant improvement ($p < 0.05$) over {#2}, while single hatching indicates significant improvement over {#1}.\xspace}

\newcommand{\arid}{Arid grid cells with no climatological precipitation are excluded.\xspace}

\begin{document}

\title{\centering
Advancing Subseasonal Forecasting with Machine Learning}

\author[1, 3]{\fnm{Hannah} \sur{Guan}}%
\equalcont{These authors contributed equally to this work.}

\author[4]{\fnm{Soukayna} \sur{Mouatadid}}%
\equalcont{These authors contributed equally to this work.}%

\author[5]{\fnm{Paulo} \sur{Orenstein}}%

\author*[6, 7]{\fnm{Judah} \sur{Cohen}}\email{jlcohen@mit.edu}

\author[2]{\fnm{Haiyu} \sur{Dong}}%

\author[2]{\fnm{Zekun} \sur{Ni}}%

\author[2]{\fnm{Jeremy} \sur{Berman}}%

\author[8]{\fnm{Genevieve} \sur{Flaspohler}}%

\author[1]{\fnm{Alex} \sur{Lu}}%

\author[9]{\fnm{Jakob} \sur{Schloer}}%

\author[9]{\fnm{Joshua} \sur{Talib}}%

\author[2]{\fnm{Jonathan A.} \sur{Weyn}}%

\author*[1]{\fnm{Lester} \sur{Mackey}}\email{lmackey@microsoft.com}

\affil[1]{\orgname{Microsoft Research}, \orgaddress{\city{Cambridge}, \state{MA}, %
\country{USA}}}

\affil[2]{\orgname{Microsoft}, \orgaddress{\city{Redmond}, \state{WA}, %
\country{USA}}}

\affil[3]{\orgname{Harvard College}, \orgaddress{\city{Cambridge}, \state{MA}, %
\country{USA}}}

\affil[4]{\orgname{University of Toronto}, \orgaddress{\city{Toronto}, \state{ON}, %
\country{Canada}}}

\affil[5]{\orgname{Instituto de Matem\'atica Pura e Aplicada}, \orgaddress{\city{Rio de Janeiro}, \state{RJ}, %
\country{Brazil}}}

\affil[6]{%
\orgname{Massachusetts Institute of Technology}, \orgaddress{\city{Cambridge}, \state{MA}, %
\country{USA}}}

\affil[7]{\orgdiv{Atmospheric and Environmental Research}, %
\orgaddress{\city{Lexington}, \state{MA}, %
\country{USA}}}

\affil[8]{\orgdiv{Rhiza Research}, \orgaddress{\city{Oakland}, \state{CA}, %
\country{USA}}}

\affil[9]{\orgname{European Centre for Medium-Range Weather Forecasts}, \orgaddress{\city{Reading}, \country{UK}}}

\abstract{
Decision-makers rely on weather forecasts to plant crops, manage wildfires, allocate water and energy, and prepare for weather extremes~\citep{white2017potential, vitart2018sub, merryfield2020current}. 
Today, such forecasts enjoy unprecedented accuracy out to two weeks thanks to steady advances in physics-based dynamical models and data-driven artificial intelligence (AI) models~\citep{bauer2015quiet,ben2024rise,bi2023accurate,lam2023learning,bodnar2025foundation,chen2023fuxi,rasp2024weatherbench}.
However, model skill drops precipitously at subseasonal timescales (2 -- 6 weeks ahead), due to compounding errors, systemic model biases, and the chaotic nature of the atmosphere \citep{vitart2012subseasonal,cohen2019s2s,chen2024machine, nathaniel2024chaosbench,richter2025scientists}.
To counter this degradation, we introduce \emph{probabilistic bias correction (PBC)}, a machine learning framework that substantially reduces systematic error by learning to correct historical probabilistic forecasts.
When applied to the leading dynamical and AI models from the European Centre for Medium-Range Weather Forecasts (ECMWF), PBC doubles the modest subseasonal skill of the AI Forecasting System~\citep{schloer2026aifssubs} and improves the skill of the operationally-debiased dynamical model~\citep{tuppi2023simultaneous} for 91\% of pressure, 92\% of temperature, and 98\% of precipitation targets. 
We designed PBC for operational deployment, and, in ECMWF’s 2025 real-time forecasting competition~\citep{loegel2025ai}, its global forecasts placed first for all weather variables and lead times, outperforming the dynamical models from six operational forecasting centers, an international dynamical multi-model ensemble, ECMWF’s AI Forecasting System, and the forecasting systems of 34 teams worldwide.
These probabilistic skill gains translate into more accurate prediction of extreme events and have the potential to improve agricultural planning, energy management, and disaster preparedness in vulnerable communities~\citep{shrader2026weather}.}

\maketitle

\section{Introduction}\label{intro}

Subseasonal forecasts---weather predictions 2 to 6 weeks ahead---lie at the heart of agricultural planning, water allocation, disaster preparedness, and energy management~\citep{white2017potential,vitart2018sub, merryfield2020current}.  
Yet accurate prediction at this timescale remains stubbornly difficult. Today, our best estimates of future weather come from dynamical models that numerically simulate how the atmosphere and oceans evolve over time~\citep{bauer2015quiet,ben2024rise} and artificial intelligence (AI) models that learn to predict the weather from historical data~\citep{bi2023accurate,lam2023learning,bodnar2025foundation}. Thanks to steady improvements in modeling, computation, and data collection over the past half-century, deterministic forecasting systems 
can now provide accurate point estimates of the weather between one and 
two weeks in advance~\citep{rasp2024weatherbench}. 
However, the two-week horizon also marks a transition in predictability: beyond this point, even small differences in initial weather conditions can lead to dramatic differences in outcomes, and a single deterministic forecast cannot capture this uncertainty~\citep{lorenz1969predictability}. 

To surmount this predictability barrier, one often turns to probabilistic forecasting, that is, predicting a distribution over future weather rather than a single point estimate~\citep{gneiting2005weather,palmer2002economic,mylne2026probability}. Probabilistic forecasting explicitly encodes forecast uncertainty and provides actionable estimates of the likelihood of high-impact extreme events. For both dynamical and AI models, probabilistic forecasts are typically generated by perturbing the inputs, parameters, or internal state of a model to produce an ensemble of possible future weather states~\citep{palmer2019ecmwf,chen2024machine,price2025probabilistic,mylne2026probability}.  While these ensembles are routinely skillful at shorter lead times (out to 15 days), operational dynamical ensembles and even state-of-the-art probabilistic AI routinely underperform the climatological average at longer subseasonal lead times, especially outside of the tropics \citep{chen2024machine}. 

A core difficulty is that leading subseasonal ensembles are generated iteratively over many time steps, and even small per-step modeling errors compound into large systematic errors over several weeks’ time \citep{nathaniel2024chaosbench}. This effect is especially pronounced for precipitation, which is sensitive to cascading inaccuracies across model components and highly variable in space and time~\citep{stephens2010dreary,bauer2015quiet}. 
Indeed, subseasonal forecasting has long been considered a “predictability desert” due to compounding model errors and complex interdependencies among the atmosphere, land, ocean, and sea ice components that evolve with different time scales~\citep{vitart2012subseasonal,richter2025scientists}.

While existing subseasonal ensembles are indeed error-prone, we show that a component of their systematic error is predictable and correctable using machine learning, allowing for significant gains in skill with minimal cost. To achieve this, we introduce a  \emph{probabilistic bias correction (PBC)} framework 
that adaptively learns to correct a model’s forecast distribution using its own historical  predictions and observations of past weather. 
PBC is compatible with any input model, and we use it to advance the state of the art in both data-driven and hybrid (AI + dynamical) subseasonal forecasting. The resulting forecasts are designed for operational deployment and, in real-time competition \cite{loegel2025ai}, outperform the best alternative dynamical, AI, and hybrid forecasting systems from forecasting centers and competitors around the globe.

\section{Results}\label{results}
\subsection{Probabilistic bias correction}

\newcommand{\figpbc}{%
\begin{figure}[htpb]
  \centering
  \opt{arxiv}{\includegraphics[width=1\textwidth]{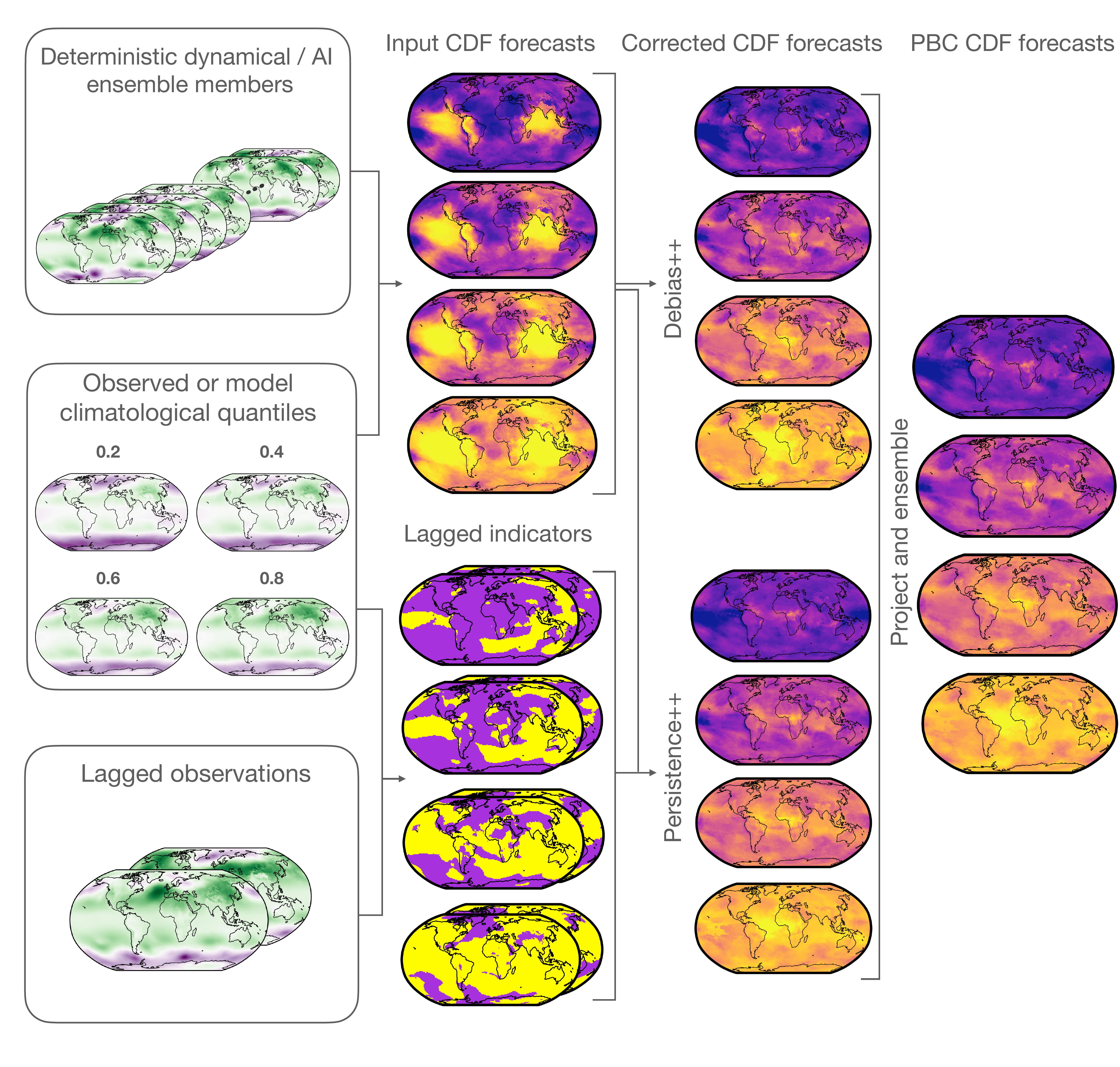}}
  \caption{
  \textbf{Probabilistic bias correction (PBC) schematic.} 
  PBC takes as input lagged observations, climatological quantiles (e.g., quintiles, terciles, or deciles), and the deterministic forecasts of any dynamical or AI model and outputs a learning-enhanced probabilistic forecast designed to correct systematic prediction errors.
  }
  \label{fig:pbc}
\end{figure}%
}

\opt{arxiv}{\figpbc}

PBC is a new machine learning framework for improving subseasonal forecasts from dynamical, AI, or hybrid prediction systems. Given a raw ensemble of deterministic forecasts, PBC outputs learning-enhanced probabilities designed to correct systematic errors while preserving the predictive signal in the underlying model.
Our approach is inspired by the deterministic bias correction framework of \citet{mouatadid2023adaptive} but operates directly on the space of probability distributions rather than at the level of raw observations. The added flexibility allows us to directly optimize not only the center of a distribution but also its spread and shape.

\cref{fig:pbc} illustrates a sample application  of PBC.  
First, an input ensemble is converted into an initial probabilistic forecast over climatological quantile bins derived from historical observations or from the model's own hindcasts. 
Next, two computationally efficient machine learning models---\debpp and \perpp---are applied in parallel to produce complementary corrections of the initial forecast. 
\debpp perturbs the forecast using location-, \mbox{date-,} and quantile-specific corrections learned by minimizing probabilistic forecasting error over adaptively-selected training periods. 
Meanwhile, \perpp blends the forecast with lagged observations and climatology to account for recent weather trends and day-of-year effects. 
Finally, the two corrections are projected onto the space of valid cumulative distribution functions (CDFs) and averaged together to form the final PBC prediction. 
See Methods for additional details.

This framework can be used to improve forecasts for any weather variable, any spatial grid, and any number of quantile bins. 
For concreteness, we adopt the following conventions of the AI Weather Quest subseasonal forecasting competition \citep{loegel2025ai} run by the European Centre for Medium-Range Weather Forecasts (ECMWF).  
First, we focus on forecasting average 2-m temperature (over land), total precipitation (over land), and average mean sea level pressure on a global 1.5° latitude-longitude grid with 5 quintile bins. 
Secondly, we evaluate subseasonal forecasts for week 3 (days 19--25) and week 4 (days 26--32) using ranked probability skill score (RPSS) \citep{weigel2007discrete}, a standard measure of forecast improvement over a climatological baseline. 
Additional details on the forecasting targets and evaluation are described in Methods.

To assess the operational utility and versatility of PBC, we apply the framework to three state-of-the-art subseasonal forecasting models: the industry-leading dynamical ensemble from ECMWF \citep{tuppi2023simultaneous}, ECMWF's subseasonal AI Forecasting System (AIFS-SUBS) \citep{schloer2026aifssubs}, and PoET, a hybrid model that post-processes the ECMWF dynamical ensemble with convolutional and transformer neural networks~\citep{benbouallegue2024improving}. 
As our measure of ground truth, we adopt the ERA5 reanalysis dataset \citep{hersbach2020era5}, which assimilates a wide range of observations to produce a global gridded record of the Earth's weather from 1940 to present, but any other target for correction could be substituted.  
Because PBC is so lightweight, we also incrementally retrain it each day, using only the reanalysis and model forecast data that would have been observable on each forecast issuance date to mimic operational deployment (Methods). 

\subsection{Correcting dynamical forecasts with PBC}

\newcommand{\figpbcecmwf}{%
\begin{figure}[htbp]
    \centering

    \opt{arxiv}{\includegraphics[width=.8\textwidth]{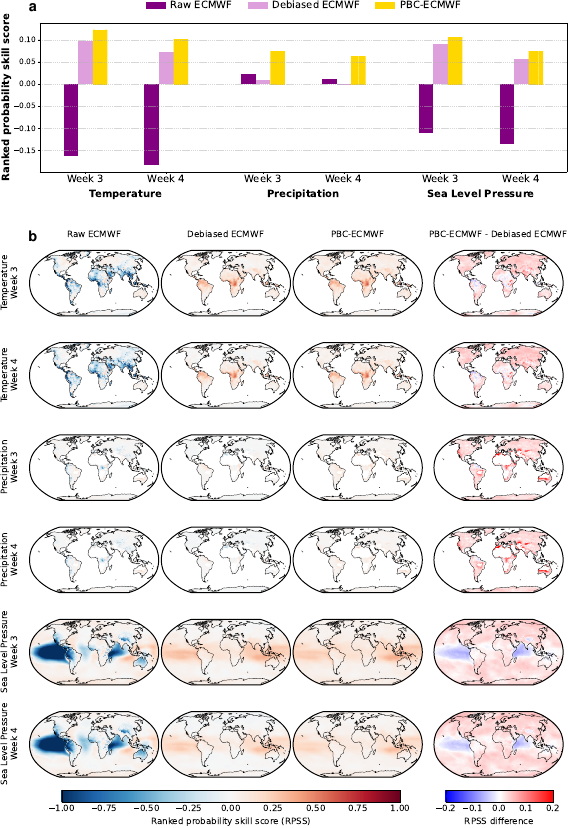}
    \vspace{\baselineskip}}
    
    \caption{\textbf{Global forecast skill of the leading dynamical model (ECMWF) and its probabilistic bias correction (PBC).}  
    (\textbf{a}) Across 
    the years 2016--2024, PBC boosts the subseasonal skill of raw ECMWF forecasts more effectively than operational debiasing, transforming low- and no-skill inputs into skillful outputs that consistently outperform climatology. 
    Negative RPSS indicates predictive performance worse than climatology,  positive RPSS indicates improved performance, and \lowercasesignificance[raw and debiased ECMWF]
    (\textbf{b}) PBC improvements are broadly distributed with 98\% of grid cells showing skill gains over debiased ECMWF for precipitation, 91--92\% for temperature, and 89--91\% for sea level pressure. 
    Arid grid cells with no climatological precipitation are excluded.
    }
    \label{fig:pbc_ecmwf_bar}
\end{figure}%
}
\opt{arxiv}{\figpbcecmwf}

\cref{fig:pbc_ecmwf_bar} summarizes the significant skill gains obtained when PBC is applied to the dynamical ensemble from ECMWF, widely considered to be the world's most accurate weather model and the gold standard for benchmarking subseasonal skill \citep{deandrade2019global,domeisen2022advances}. 
For each target variable and lead time, \cref{fig:pbc_ecmwf_bar}a reports global RPSS over the years 2016--2024, while \cref{fig:pbc_ecmwf_bar}b displays the spatial RPSS distribution over the same time period. 
Negative RPSS indicates worse predictive performance than the climatological baseline, while positive RPSS indicates improved performance. 
PBC consistently improves the low-scoring raw ECMWF ensemble, 
tripling its positive precipitation skill (0.023 RPSS) 
and converting its negative-skill temperature 
and mean sea level pressure forecasts into skillful predictions that consistently outperform climatology (\cref{fig:pbc_ecmwf_bar}a). 
Analogous gains are observed when skill is stratified by season (Extended Data \cref{fig:pbc_ecmwf_seasonal}) or by region (Extended Data \cref{fig:pbc_ecmwf_regional}), and the improvements are accompanied by a pronounced reduction in systematic model bias across the globe (Extended Data \cref{fig:bias_pr,,fig:bias_tas,,fig:bias_mslp}). 

To dampen the systemic effects of model bias, forecasting centers routinely debias their operational forecasts by comparing each deterministic prediction to quantiles derived from model hindcasts rather than observations \citep{SUBS-M-climate}.  
Given the same deterministic predictions as input, PBC consistently outperforms this operational debiasing protocol, replacing the near-zero precipitation skill of debiased ECMWF with consistently positive RPSS and boosting the positive temperature and pressure RPSS of debiased ECMWF by 26--37\% and 16--32\% (over baseline skills of 0.07--0.10 and 0.06--0.11) respectively (\cref{fig:pbc_ecmwf_bar}a). 
The PBC gains over debiased ECMWF are also broadly distributed with 98\% of grid cells showing skill gains for precipitation, 91--92\% for temperature, and 89--91\% for sea level pressure (\cref{fig:pbc_ecmwf_bar}b).
These results demonstrate that PBC can transform unskillful inputs into reliably skillful outputs and serve as an improved drop-in replacement for existing operational debiasing protocols.

In fact, PBC can boost the skill of its dynamical input beyond the levels of existing AI-based subseasonal models. 
For example, \citet{chen2024machine} recently introduced FuXi-S2S, a deep learning model that outperforms ECMWF's dynamical ensemble at subseasonal lead times. 
Across the globe and the test years 2017--2021, PBC-ECMWF displays a pronounced increase in skill over FuXi-S2S for each variable and lead time, 
with 99--100\% of grid cells showing skill gains for temperature, 97--99\% for precipitation, and 96--100\% for mean sea level pressure (Extended Data \cref{fig:fuxi}). 

Decomposing PBC into its component parts (Extended Data \cref{fig:pbc_breakdown}a) yields four additional insights into its skill gains.
First, \debpp and \perpp each significantly improve ($p <0.05$) upon raw ECMWF for all targets.
Second, \perpp alone significantly improves ($p <0.05$) upon debiased ECMWF for each target save week 4 pressure, while \debpp provides smaller improvements.
Third, projection minimally impacts skill as \debpp and \perpp forecasts are typically valid CDFs.
Finally, ensembling the complementary \perpp and \debpp forecasts offers robust improvements across all variables and lead times.

\clearpage
\subsection{Correcting AI forecasts with PBC}

\newcommand{\figpbcai}{%
\begin{figure}[htbp]
    \centering

    \opt{arxiv}{\includegraphics[width=\textwidth]{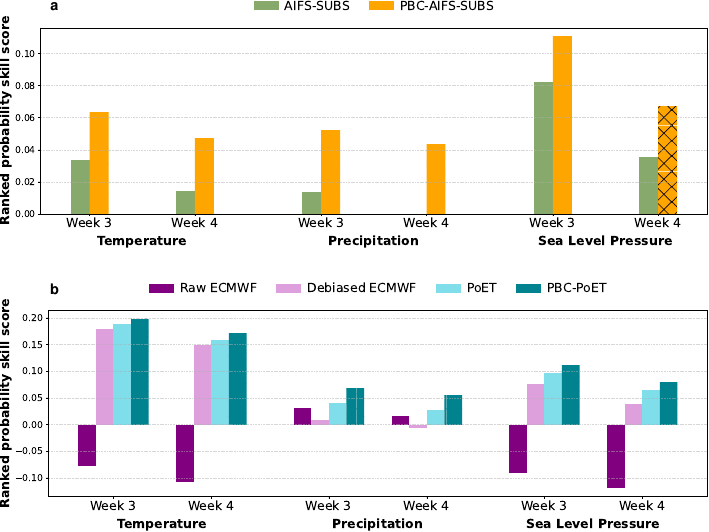}
    \vspace{0\baselineskip}}
    \caption{\textbf{Global forecast skill of AI models and their probabilistic bias corrections (PBC).} 
    (\textbf{a}) In 2025, PBC consistently enhances the subseasonal skill of ECMWF's AI Forecasting System (AIFS-SUBS), 
    doubling its average skill (0.03) across all variables and lead times.
    (\textbf{b}) In 2024, the hybrid (AI + dynamical) model PoET already improves the skill of its dynamical input, ECMWF. Nevertheless, PBC further enhances skill for every variable and lead time with gains of 
    68--98\% for precipitation (over baseline skills of 0.03--0.04), 
    15--23\% for mean sea level pressure (over baseline skills of 0.08--0.10),
    and
    5--7\% for temperature (over baseline skills of 0.16--0.19).
    \significance}
    \label{fig:pbc_ai_rpss}
\end{figure}
}
\opt{arxiv}{\figpbcai}

PBC can also be used to improve the skill of AI models. 
For example, \cref{fig:pbc_ai_rpss}a shows the substantial skill gains obtained when PBC is applied to ECMWF's state-of-the-art subseasonal AI Forecasting System (AIFS-SUBS). The AIFS is a deep learning model that exhibits greater subseasonal skill than debiased ECMWF for the challenging task of precipitation forecasting and comparable skill for temperature and mean sea level pressure prediction \cite{schloer2026aifssubs}. 
Across the globe and the 2025 test year, PBC consistently amplifies AIFS-SUBS skill, doubling its mean RPSS (0.03) across all variables and lead times.

We also apply PBC to a hybrid (AI + dynamical) forecasting system, PoET.  
PoET takes as input a dynamical ensemble (in this case, ECMWF) and outputs a new ensemble with all members simultaneously corrected by a transformer embedded within a convolutional neural network \citep{benbouallegue2024improving}. 
PoET serves as challenging input for PBC as its ensemble has already been corrected by AI and is substantially more skillful than its raw ECMWF input. 
Moreover, PoET has been shown to outperform other statistical and AI-based post-processing techniques including member-by-member model output statistics~\citep{van2015ensemble,vannitsem2020statistical}, ensemble model output statistics~\citep{gneiting2005calibrated}, LeNet neural network post-processing~\citep{li2022convolutional}, and pure transformer-based post-processing~\citep{finn2021self}. 
Nevertheless, PBC enhances PoET skill for every variable and lead time with RPSS gains of 
68--98\% for precipitation (over baseline skills of 0.03--0.04), 
15--23\% for mean sea level pressure (over baseline skills of 0.08--0.10),
and
5--7\% for temperature (over baseline skills of 0.16--0.19) in 2024 (\cref{fig:pbc_ai_rpss}b).

A decomposition of PBC into its component parts (Extended Data \cref{fig:pbc_breakdown}b) reveals three distinct features contributing to its improvements.
First, \debpp alone improves upon its PoET input for each task, highlighting the value of direct CDF optimization 
over correction only in deterministic forecast space~\citep{mouatadid2023adaptive,benbouallegue2024improving,van2015ensemble,vannitsem2020statistical,finn2021self}.
Second, \perpp delivers even greater skill for precipitation, highlighting the value of 
integrating recent observations
over a correction that ignores this source of predictability~\citep{benbouallegue2024improving,van2015ensemble,vannitsem2020statistical,gneiting2005calibrated,finn2021self,scheuerer2020using}. 
Third, ensembling the two complementary corrections robustly captures the strengths of each, highlighting the value of model merging over monolithic correction~\citep{benbouallegue2024improving,van2015ensemble,vannitsem2020statistical,gneiting2005calibrated,finn2021self,li2022convolutional,scheuerer2020using}.

\subsection{AI Weather Quest forecasting competition}
\newcommand{\figaiwq}{%
\begin{figure}[H]
    \centering
    \opt{arxiv}{\includegraphics[trim=.25cm 0 .25cm 0,clip,width=\linewidth]{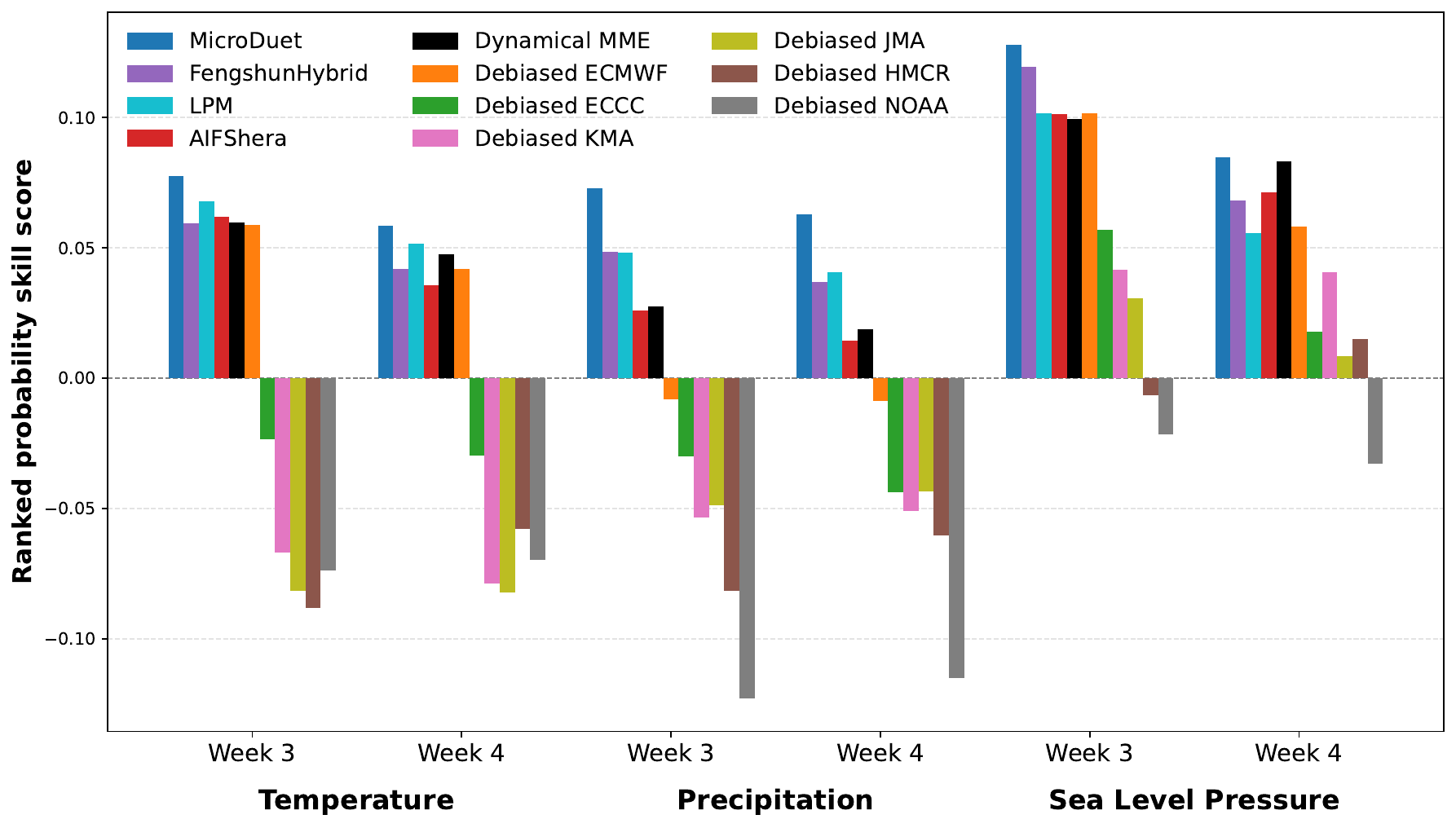}} 
    \caption{\textbf{AI Weather Quest real-time global forecast skill.} 
    MicroDuet, an ensemble of PBC-ECMWF and PBC-PoET, placed first in the  2025 September-October-November AI Weather Quest competition for every variable and lead time, outperforming the debiased dynamical models of six government agencies (ECMWF, NOAA, ECCC, HMCR, KMA, and JMA), their dynamical multi-model ensemble (Dynamical~MME), three ECMWF AI models (including their top-performing AIFShera model), and the forecasting systems of 34 teams worldwide (including the top competitors, FengshunHybrid and LPM).}
    \label{fig:aiwq}
\end{figure}%
}
\opt{arxiv}{\figaiwq}

In 2025, ECMWF launched the AI Weather Quest, an international competition to benchmark and improve the state of operational subseasonal forecasting \citep{loegel2025ai}. 
Each week, contestants submit real-time probabilistic forecasts for weather around the globe, and, at the end of each season, models with the highest week 3 and week 4 skills are recognized as winners. 
To benchmark the real-time operational utility of PBC, we entered an ensemble of PBC-ECMWF and PBC-PoET into the competition under the name MicroDuet. 
In the inaugural September-October-November season, MicroDuet placed first for all weather variables and lead times, outperforming the debiased dynamical forecasts of six operational forecasting centers (ECMWF, NOAA, ECCC, HMCR, KMA, and JMA) \citep{vitart2025wwrp}, a multi-model ensemble of these six dynamical forecasts, three versions of ECMWF's AI Forecasting System, and the forecasting systems of 34 teams worldwide (\cref{fig:aiwq}). 

The MicroDuet skill gains are especially pronounced for precipitation, as the debiased models of all six operational forecasting centers exhibit \emph{negative} skill, performing worse than the climatological average. 
The Dynamical MME recovers positive skill by aggregating across all six forecasting centers, but, with skills of 0.06--0.07, MicroDuet alone is twice as skillful in week 3 and thrice as skillful in week 4. 
MicroDuet also placed first for both lead times in the December-January-February season of the Quest and, across $37$ teams, was the only model to outperform Dynamical MME for all variables \citep{ecmwf2026aiweatherquestleaderboard}. 
Across both seasons, MicroDuet was also recognized as the only submission to significantly outperform debiased ECMWF ($p<0.05$) for all variables and lead times  \citep{talib2026aiweatherquest}. 

In November 2025, MicroDuet’s week 4 temperature forecast of a mid-December cold air outbreak in the Eastern United States was shared in real-time \citep{oliveira2025northeast} and verified as an extreme cold event \citep{atienza2025polar} that was missed by debiased ECMWF (Extended Data \cref{fig:aiwq_extreme}). This demonstrates the potential of PBC to improve not only general forecast quality but also the early detection of impactful extreme weather events. We next assess the skill of PBC when trained to target such extreme events.

\subsection{Forecasting extreme events}

The early identification of extreme weather is an important application of subseasonal forecasting that supports improved preparedness for hazards such as floods, heatwaves, droughts, storms, and cold snaps.  
\cref{fig:extremes3} summarizes the significant gains in extreme forecasting skill when PBC is applied to the dynamical ensemble from ECMWF. 
Following prior work~\citep{diffenbaugh2005fine}, we define extreme highs as events verifying above the 95th percentile and extreme lows as events verifying below the 5th percentile. 
We then train PBC to target semidecile (5\% quantile) increments rather than quintiles and measure skill using the Brier skill score (BSS)~\citep{weigel2007discrete}, a standard measure of improvement over a climatological baseline for a binary event.

For each target variable and extreme, \cref{fig:extremes3}a reports week 3 global BSS over the years 2016--2024, while \cref{fig:extremes3}b displays the spatial BSS distribution over the same time period. 
For extreme precipitation and all extreme lows, PBC again transforms its low- and no-skill inputs into skillful outputs that consistently outperform climatology. For extreme highs, PBC boosts the forecasting skill of operationally debiased ECMWF by 51\% (over a baseline skill of 0.07) for temperature and 39\% (over a baseline skill of 0.05) for pressure (\cref{fig:extremes3}a).
The PBC improvements over debiased ECMWF are also broadly distributed with 96\% of grid cells showing skill gains for precipitation, 89\% for sea level pressure, and 85\% for  temperature (\cref{fig:extremes3}b).
Comparable improvements are also observed in week 4 (Extended Data \cref{fig:extremes4}).

To evaluate whether these gains translate into improved hazard prediction, \cref{fig:extremes3}c  displays flood forecasting skill across all flood events catalogued by the Global Disaster Alert and Coordination System (GDACS) \citep{masante2025multi} in the years 2016--2026.
For each event, flood forecasting skill is quantified by the BSS for the top precipitation semidecile across the $20^\circ\times 20^\circ$ bounding box centered on the flood centroid. 
Given the same raw ECMWF forecasts as input, 
PBC boosts flood forecast skill by 76--92\%, while debiased ECMWF dramatically reduces it, driving 0.04--0.06 baseline skills to near zero. 
GDACS additionally categorizes each flood based on its humanitarian impact and labels a flood severe (Orange or Red) if it results in more than 100 deaths or 80,000 displacements \citep{masante2025multi}. 
For these high-impact events, operational debiasing only degrades or matches the 0.04--0.07 BSS of raw ECMWF, while PBC enhances its severe flood forecasting skill by 86--150\%. 

\newcommand{\figextremesthree}{%
\begin{figure}[H]
    \centering

    \opt{arxiv}{\includegraphics[width=\linewidth]{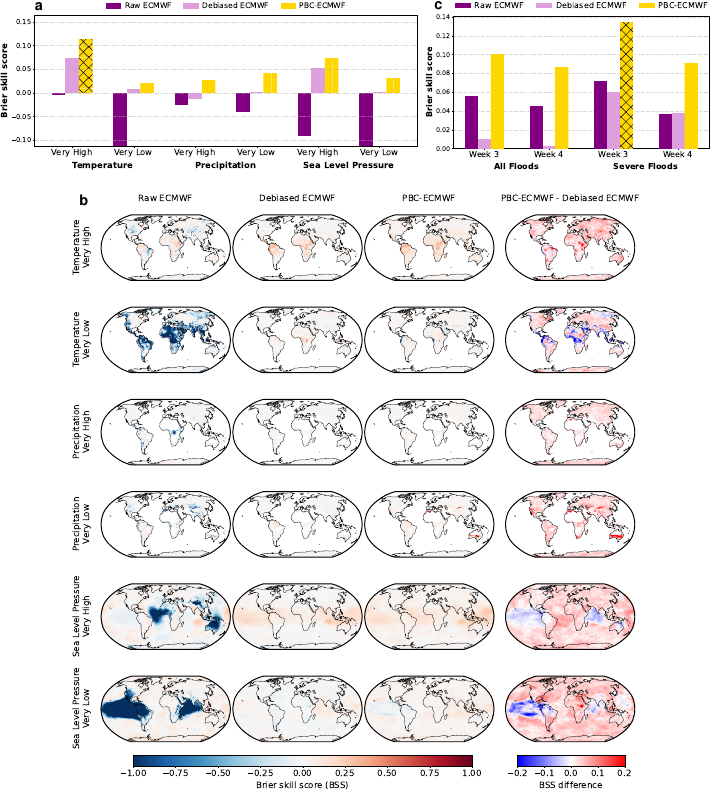}}

    \caption{\textbf{Forecasting 
    extreme weather
    with the leading dynamical model (ECMWF) and its probabilistic bias correction (PBC).} (\textbf{a}, \textbf{b}) Across the globe and the years 2016--2024, PBC boosts ECMWF week 3 extreme forecast skill more effectively than operational debiasing with 96\% of grid cells showing skill gains for precipitation, 89\% for sea level pressure, and 85\% for temperature. 
    Arid grid cells with no climatological precipitation are excluded.
    (\textbf{c}) Across all 2016--2026 flood events catalogued by the Global Disaster Awareness and Coordination System (GDACS) and severe GDACS floods with over $100$ deaths or $80,000$ individuals displaced, 
    operational debiasing routinely \emph{reduces} raw ECMWF flood forecast skill, while
    PBC delivers pronounced improvements. 
    \significance[raw and debiased ECMWF]}
    \label{fig:extremes3}
\end{figure}
}
\opt{arxiv}{\figextremesthree}

\section{Discussion}\label{discussion}

Our results indicate that probabilistic post-processing with machine learning can substantially boost the quality and utility of subseasonal forecasts. The probabilistic bias correction (PBC) framework achieves these goals by satisfying four key properties. First, across the globe, PBC consistently improves the world’s leading dynamical model from ECMWF, yielding a new state of the art for hybrid (AI + dynamical) subseasonal forecasting. Second, PBC consistently improves leading subseasonal AI models, establishing a new state of the art for fully data-driven subseasonal forecasting. 
Third, PBC forecasts are suitable for operational deployment and, in real-time competition, consistently outperform the best alternative dynamical, AI, and hybrid forecasting systems from around the world.
Fourth, PBC consistently improves the early detection of extreme and high-impact weather events. As climate change is predicted to increase the spatial and temporal occurrence of extreme weather, improvements in early warning capability have increasing potential to reduce preventable morbidity and mortality~\citep{shrader2026weather}.

PBC targets predictable errors rather than attempting to relearn full atmospheric dynamics. 
Hence, it should be viewed as an effective post-processing tool rather than a replacement for input model improvement. Fortunately, the PBC framework is also inherently adaptive: as dynamical and AI models are upgraded with improved physical representation, process representations, computational capacity, or training data, PBC can be retrained to ingest and correct these higher-quality inputs. Because PBC is also inexpensive when compared with the generation of large-scale dynamical or AI ensembles, the framework provides a low-overhead strategy for meteorological centers, including those in resource-constrained settings, to improve their existing forecast products. 

With its subseasonal lead times, PBC also targets a critical gap in the weather forecasting spectrum: at longer leads, many high-impact events are not yet predictable, while, at shorter leads, many warnings are no longer actionable. 
For example, long-lead seasonal forecasts failed to predict the disastrous 2018 Kenyan floods, but subseasonal rainfall forecasts provided a clear warning while preparation was still possible \citep{kilavi2018extreme}. 
A growing body of evidence demonstrates the value of skillful subseasonal forecasts for disaster response, energy production, healthcare, and agriculture \citep{white2017potential, vitart2018sub, merryfield2020current}, especially in semi-arid regions in Eastern and Southern Africa, where agriculture remains highly susceptible to out-of-season rains \citep{kolstad2025skilful,mwangi2024variability}.  However, this evidence often takes the form of isolated case studies leveraging custom post-processing to extract signal from raw model outputs \citep{turner2025investigating}. 
PBC represents a significant advance for subseasonal forecasting as its low-cost skill gains generalize across target variables, input models, and regions. 
Through the accompanying release of our open-source code, we aim to provide a standardized tool for the community to bridge the gap between raw forecasts and actionable intelligence at subseasonal horizons and beyond.

\bibliography{refs}%

\opt{nature}{
\figpbc
\figpbcecmwf
\figpbcai
\figaiwq
\figextremesthree
}
\section{Methods}\label{methods}

\subsection{Probabilistic subseasonal forecasting}
We begin with a formal description of the probabilistic subseasonal forecasting problem, following the conventions of ECMWF's 2025 AI Weather Quest competition \citep{loegel2025ai}. 
For each week-long period beginning on a date $t$, let $\gt_t \in \reals^G$ represent a vector of ground-truth weather measurements $y_{t,g}$ for each grid cell $g$ in a global latitude-longitude grid. We used 1979-2026 ERA5 global reanalysis \citep{hersbach2020era5} regridded to  $1.5^\circ\times1.5^\circ$ latitude-longitude resolution as our measure of ground truth and focus on predicting average 2-meter temperature, total precipitation, and average mean sea level pressure across each week-long period. Both temperature and precipitation are evaluated only over land-dominated regions (i.e., grid cells with at least $50\%$ land coverage).

For each grid cell $g$, our aim is to predict a distribution over weather at time $t$, expressed as a cumulative distribution function (CDF) $F_{t,g}$
over $K=5$ historical climatological quantile bins. To this end, we set $\qobs_{t,g}(K)=\infty$ and, for each $k < K$, define the $k$-th climatological quintile $\qobs_{t,g}(k)$ for $(t,g)$ as the $\frac{k}{K}$-th quantile of the observations $y_{s,g}$ falling $0$, $2$, or $4$ days away from the target month-day combination in each of the past $20$ years. For each $k$, the probabilistic forecast $F_{t,g}(k)$ represents the predicted probability that the measurement $y_{t,g}$ will fall below the $k$-th quintile threshold $\qobs_{t,g}(k)$. 

Given a test set $\testset$ of target dates, we measure 
the global quality of these probabilistic forecasts via the ranked probability skill score (RPSS) \citep{weigel2007discrete},
\begin{align}
    \rpss = 1-\frac{\sum_{t\in\testset}\sum_{g=1}^G w_g\,\rps_{t,g}}{\sum_{t\in\testset}\sum_{g=1}^G w_g\,\rpsc_{t,g}} 
    \sstext{for}
    \rps_{t,g} = \sum_{k=1}^K(F_{t,g}(k) - O_{t,g}(k))^2 
    \sstext{and}
    \rpsc_{t,g} = \sum_{k=1}^K\left(\!\frac{k}{K} - O_{t,g}(k)\!\right)^{\!2}\!,
\end{align}
where $O_{t,g}(k) = \indic{y_{t,g} < \qobs_{t,g}(k)}$ is the observed indicator of whether $y_{t,g}$ falls below the $k$-th quintile threshold, 
$w_g = \cos(\latitude(g))$ adjusts for the relative size of each grid cell, 
$\rps_{t,g}$ is termed the ranked probability score (RPS) of the forecast, and $\rpsc_{t,g}$ is the RPS of a baseline climatological forecast that assigns equal probability to each quintile bin. 
For precipitation, arid grid cells $g$ with no climatological precipitation (i.e., with $q_{t,g}(K-1)=0$) are excluded from this evaluation.

Similarly, for a given grid cell $g$, we measure forecast quality over a test set $\testset$ via the spatial RPSS
\begin{align}
\srpss_g 
    =
1 - \frac{\sum_{t\in\testset} \rps_{t,g}}{\sum_{t\in\testset} \rpsc_{t,g}}.
\end{align}
Throughout, we evaluate RPSS at two subseasonal forecasting horizons, days 19--25 (week $3$) and days 26--32 (week $4$).
Unless otherwise noted, we use all non-leap-day Monday and Friday target dates in the years 2016--2024 as our default test set.

\subsection{Baselines}\label{sec:baselines}
\cref{tab:baselines} provides a summary of the dynamical, AI, hybrid (dynamical + AI), and post-processing baselines compared with PBC in this work. Here we detail the processing steps taken to evaluate each baseline. 
\subsubsection{Raw ECMWF}
At each subseasonal lead time $\ell$, ECMWF's operational dynamical model \citep{tuppi2023simultaneous} outputs an ensemble of deterministic forecasts,  $(\hat{y}_{m,\ell,t,g})_{m=1}^M$,  each representing a possible future weather state. 
For each target weather variable and subseasonal horizon, we downloaded all weekly-aggregated ECMWF ensemble forecasts, spanning the years 2015--2026,  from the Subseasonal to Seasonal (S2S) Prediction Project Database \citepmethods{vitart2017subseasonal} via the International Research Institute for Climate and Society Data Library (IRIDL) \citepmethods{ecmwfdata2026}. 
Forecasts with $M=51$ members ($1$ control and $50$ perturbed) were issued every Monday and Thursday from May 14, 2015 -- June 26, 2023, and forecasts with $M=101$ members were issued daily starting on June 28, 2023. 
We converted each ensemble into a CDF forecast $F_{\ell,t,g}(k) = \frac{1}{M}\sum_{m=1}^M \indic{\hat{y}_{m,\ell,t,g} < \qobs_{t,g}(k)}$ using the climatological quintile thresholds $\qobs_{t,g}$. 

\subsubsection{Debiased ECMWF}

To form the debiased ECMWF baseline, we followed ECMWF's operational post-processing protocol \citep{SUBS-M-climate} for each weather variable and subseasonal horizon. 
First, we downloaded all weekly-aggregated ECMWF ensemble hindcasts from the S2S database \citepmethods{vitart2017subseasonal} via the IRIDL \citepmethods{ecmwfdata2026}. 
Each ensemble hindcast consists of $M=11$ members, $(\hat{y}_{m,\ell,\delta,t,g})_{m=1}^M$, indexed by a forecast target date $t$ and the number of years $\delta\in\{1,\dots,20\}$ between the forecast and corresponding hindcast target dates. 
Next, for each forecast target date $t$, we identified the set $\mathcal{S}_t$ of model climatology dates: the closest date no later than $t$ with available hindcasts along with the next 4 and previous 4 available dates. 
We then formed model-based quintile thresholds for each forecast target date $t$, lead time $\ell$, and grid cell $g$ by setting 
$\qecmwf_{\ell,t,g}(K)=\infty$ and, for each $k < K$, defining the $k$-th model quintile $\qecmwf_{\ell,t,g}(k)$ as the $\frac{k}{K}$-th quantile of all hindcasts $\hat{y}_{m,\ell,\delta,s,g}$ with $s \in \mathcal{S}_t$.
Finally, we constructed the debiased ECMWF CDF forecast $F_{\ell,t,g}(k) = \frac{1}{M}\sum_{m=1}^M \indic{\hat{y}_{m,\ell,t,g} < \qecmwf_{\ell,t,g}(k)}$.

\subsubsection{FuXi-S2S}
For each target weather variable and subseasonal horizon, we downloaded 51-member ensemble forecasts, spanning the test years 2017--2021, and 51-member ensemble hindcasts, spanning the training years 2002--2016, from the official FuXi-S2S forecast dataset \citepmethods{zhongfuxi2025}. Forecasts were subsequently aggregated to a weekly resolution, with precipitation accumulated and temperature and mean sea level pressure averaged.
Following \citet{chen2024machine}, we form model-based quintile thresholds  
for each forecast target date $t$, lead time $\ell$, and grid cell $g$ by setting 
$\qfuxi_{\ell,t,g}(K)=\infty$ and, for each $k < K$, defining the $k$-th model quintile $\qfuxi_{\ell,t,g}(k)$ as the $\frac{k}{K}$-th quantile of all hindcasts matching the target month-day combination. Finally, we constructed the CDF forecast $F_{\ell,t,g}(k) = \frac{1}{M}\sum_{m=1}^M \indic{\hat{y}_{m,\ell,t,g} < \qfuxi_{\ell,t,g}(k)}$ from each FuXi-S2S ensemble forecast $(\hat{y}_{m,\ell,t,g})_{m=1}^{M}$ with $M=51$. 

Since FuXi-S2S and ECMWF historically issued their forecasts on different schedules, we evaluated all available forecasts from FuXi-S2S and PBC-ECMWF with target dates in the period 2017--2021. This corresponds to all Monday and Friday target dates for PBC-ECMWF and, for FuXi-S2S, all Tuesday and Friday target dates in 2017, 
all Wednesday and Saturday target dates in 2018,
all Thursday and Sunday target dates in 2019, 
all Monday and Friday target dates from 2020-01-03 to 2020-02-24, 
all Tuesday and Saturday target dates from 2020-02-29 to 2020-12-29, 
and all Wednesday and Sunday target dates 
from 2021-01-03 to 2021-12-26.

\subsubsection{AIFS-SUBS}

For each target weather variable, subseasonal horizon, and odd issuance date in 2025 (save for the 31st of any month), we generated weekly-aggregated $100$-member ensemble forecasts using the AIFS-SUBS model \citep{schloer2026aifssubs}. AIFS-SUBS follows the architecture introduced in \citetmethods{lang2024aifs}, combining a transformer-based graph neural network encoder and decoder with a sliding-window transformer processor in which attention is computed along spiral longitudinal bands. Ensemble members are generated by conditioning the processor on independently sampled noise, injected via conditional layer normalization in each processor layer, and the model is trained with the fair continuous ranked probability score loss computed over $4$ ensemble members against ERA5 targets \citepmethods{lang2024aifscrps}. 
AIFS-SUBS takes two consecutive states ($t-24$h, $t$) as input and predicts the atmospheric state at $t+24$h. Inputs and outputs are represented on the reduced Gaussian o96 grid (approximately $1^\circ$ resolution). The variant of AIFS-SUBS used in this work includes a representation of the stratosphere and was trained exclusively on ERA5 over the periods 1979--2006 and 2012--2024, leaving a 5-year validation period. 

Following \citet{schloer2026aifssubs}, 
for each forecast target date $t$, lead time $\ell$, and grid cell $g$, we generated $10$-member ensemble hindcasts for the corresponding month-day combination in each of the prior 20 years, set
$\qaifs_{\ell,t,g}(K)=\infty$, and, for each $k < K$, defined the $k$-th model quintile $\qaifs_{\ell,t,g}(k)$ as the $\frac{k}{K}$-th quantile of the aforementioned hindcasts.  
Finally, we constructed the CDF forecast $F_{\ell,t,g}(k) = \frac{1}{M}\sum_{m=1}^M \indic{\hat{y}_{m,\ell,t,g} < \qaifs_{\ell,t,g}(k)}$ from each AIFS ensemble forecast $(\hat{y}_{m,\ell,t,g})_{m=1}^{M}$ with $M=100$. 

\subsubsection{PoET}

The Post-processing Ensembles with Transformers (PoET) method \citep{benbouallegue2024improving} is a member-by-member post-processing technique that has been shown to significantly improve the medium-range weather forecasting skill of ECMWF dynamical forecasts and to outperform other statistical and AI-based post-processing techniques including the member-by-member model output statistics post-processing technique \citep{van2015ensemble} used operationally at the Royal Meteorological Institute of Belgium \citep{vannitsem2020statistical}, ensemble model output statistics~\citep{gneiting2005calibrated}, LeNet neural network post-processing~\citep{li2022convolutional}, and pure transformer-based post-processing~\citep{finn2021self}. 
PoET includes a self-attention mechanism across the ensemble member dimension within its encoder-decoder convolutional U-Net architecture, allowing the model to use context from the full input forecast distribution to minimize probabilistic forecasting error. 
We extended PoET to subseasonal horizons by making a few minor modifications to its data processing pipeline and architecture. 

First, we applied PoET to weekly aggregates of weather variables rather than the instantaneous values appropriate for medium-range forecasting. 
As input features, we used all available variables from the S2S database \citepmethods{vitart2017subseasonal} and a collection of 36 constant fields from ERA5 \citep{hersbach2020era5}, including orography, a land-sea mask, and soil and vegetation types. 
Following common practice for feature embeddings, metadata, including the constant fields, lead time, and weekly-averaged top-of-atmosphere incoming solar radiation, were embedded with a learned convolutional layer and used to update the highest-level features additively. 
To accommodate the larger set of input and output features, the model size was increased to approximately 27 million trainable parameters by increasing the number of latent features and the number of ensemble transformer layers.

PoET is configured by default to predict multiple targets, including geopotential heights, wind, and humidity on a few pressure levels and surface winds and humidity, in addition to the target variables temperature, mean sea level pressure, and precipitation. 
While predicting multiple targets improves consistency between variables, we found that the single-target model produced higher-quality outputs for precipitation on our validation set. 
Hence, the final results reported for PoET are from the multi-target model for temperature and pressure and from the single-target model for precipitation. 
The PoET models were developed using the 20 years of 10-member ECMWF hindcasts issued in 2024, with the hindcasts for the years 2004--2022 used for training and the 2023 hindcasts used for validation. 

For each target weather variable, subseasonal horizon, and ECMWF forecast issuance date in 2024 (every Monday and Thursday through November 21 and every odd date thereafter), we generated weekly-aggregated ensemble forecasts, $(\hat{y}_{m,\ell,t,g})_{m=1}^M$, by applying PoET to the $M=100$ perturbed ECMWF ensemble members. We then converted each ensemble into a CDF forecast $F_{\ell,t,g}(k) = \frac{1}{M}\sum_{m=1}^M \indic{\hat{y}_{m,\ell,t,g} < \qobs_{t,g}(k)}$ using the climatological quintile thresholds $\qobs_{t,g}$.

\subsection{Probabilistic bias correction (PBC)}\label{sec:pbc-methods}

PBC (\cref{fig:pbc}) corrects the systematic errors in probabilistic forecasts by training and applying two complementary machine learning corrections (\debpp and \perpp), projecting the output of each onto the space of valid distributions, and finally averaging the resulting distributions. 
For a given weather variable, target date $\tstar$, and lead time $\lstar$, each machine learning algorithm is given access to a model's uncorrected CDF forecasts $\fcst_{\lstar, \tstar} = (F_{\lstar, \tstar, g})_{g=1}^G$ for the target, a dataset of historical CDF forecasts $\fcst_{\ell, t} = (F_{\ell, t, g})_{g=1}^G$ from the model with training set target dates $t\in\trainset$ and lead times $\ell\in\leads$, and observed cumulative quantile bin indicators $\ind_t = (O_{t,g})_{g=1}^G$ for each $t\in\trainset$. 
To mimic operational deployment, each algorithm is only given access to training observations that would have been fully observable $\lstar-1$ days prior to the target date $\tstar$. 

\subsubsection{\debpp}

For each quantile bin $k$, \debpp (\cref{alg:debpp}) performs probabilistic forecast correction in four steps. 
First, to account for heterogeneity in forecast errors over time, \debpp adaptively selects a span $s$ of observations around the target day of year for probabilistic debiasing and a range $\dstar$ of forecast issuance dates for ensembling. 
Second, \debpp forms ensemble mean probabilistic forecasts $\overline{\fcst}_t(k)$ by averaging its input CDF forecasts over the selected range of issuance dates. 
Third, to directly counteract systematic errors, \debpp corrects the target CDF forecast $\overline{\fcst}_{\tstar}(k)$ by subtracting the probabilistic prediction bias---the average forecast value $\overline{\fcst}_t(k)$ minus the average observed value $\ind_t(k)$---over the selected debiasing window across the past $Y$ years. 
Finally, the results are projected into the valid probability range $[0,1]$. 

In every application of \debpp, we set $Y=20$, selected the configuration $(s,\dstar)\in\{(14,1),(28,1),(35,1)\}$ that minimized mean RPS over the 3 years preceding the target date $\tstar$, and converted each available ensemble forecast $(\hat{y}_{m,\ell,t,g})_{m=1}^M$ and hindcast $(\hat{y}_{m,\ell,\delta,t,g})_{m=1}^M$ into an input CDF forecast using the observed climatological quintile thresholds $q_{t,g}$:
\begin{align}
F_{\ell,t,g}(k) = \frac{1}{M}\sum_{m=1}^M \indic{\hat{y}_{m,\ell,t,g} < q_{t,g}(k)}
\qtext{and}
F_{\ell,t-\delta\,\textup{years},g}(k) = \frac{1}{M}\sum_{m=1}^M \indic{\hat{y}_{m,\ell,\delta,t,g} < q_{t-\delta\,\textup{years},g}(k)}.
\end{align}

\begin{algorithm}%
  \caption{\debpp}
  \label{alg:debpp}
  \begin{algorithmic}
  
  	\INPUT target date $\tstar$; 
    quantile bin $k$; 
  	lead time $\lstar$; 
    span $s$; 
  	\# issuance dates $\dstar$; 
    \# training years $Y$; 
    training set of observed cumulative quantile bin indicators $(\ind_t)_{t\in \trainset}$ and cumulative probability forecasts $(\fcst_{\ell,t})_{\ell \in\leads, t\in \trainset\cup\{\tstar\}}$
    \INIT days per year $D = 365.242199$
    \STATE $\mc{S} 
	= \{t \in \trainset:
	\texttt{year\_diff}\coloneqq \floor{\frac{\tstar-t}{D}} \leq Y
	\text{ and } 
	\texttt{day\_diff}\coloneqq \frac{365}{2} -|\floor{(\tstar-t)\mod D}- \frac{365}{2} | \leq s 
	\}$\\[1pt]
    \STATE // Form ensemble mean probabilistic forecasts across issuance dates 
    \FOR{target and training dates $t \in \mc{S}\cup \{\tstar\}$}
    \STATE
    $
    \bar{\fcst}_t(k)
    = \textup{mean}((\fcst_{\lstar-d+1,t}(k))_{1 \leq d \leq \dstar})
    $ \\[2pt]
    \ENDFOR 
    \OUTPUT $\clip\left(\bar{\fcst}_{\tstar}(k) + \textup{mean}((\ind_t(k) - \bar{\fcst}_{t}(k))_{t\in \mc{S}}), [0,1]\right)$
  \end{algorithmic}
\end{algorithm}

\subsubsection{\perpp}
\label{sec:perpp}

For each quantile bin $k$ and grid cell $g$, \perpp (\cref{alg:perpp}) accounts for both recent weather trends and climatological shifts by regressing the observed indicator $O_{t,g}(k)$ onto the historical model forecast $F_{\lstar,t,g}(k)$, two recent and fully-observable lagged indicators $O_{t-\lstar+6,g}(k)$ and $O_{t-2\lstar+5,g}(k)$, and a probabilistic climatology formed from the preceding 20 years of observed indicators with matching month and day. In the climatology computation, any missing indicators are imputed with the prior value $\frac{k}{K}$. 
The learned regression weights are applied to the target forecast $F_{\lstar,\tstar,g}(k)$, lagged observations, and climatology to produce a corrected output that is then projected into the valid probability range $[0,1]$. 

When applying \perpp to PoET forecasts or ECMWF precipitation forecasts, we again converted each available ensemble forecast and hindcast into an input CDF forecast using the observed climatological quintile thresholds. 
In all other applications of \perpp, we used model-based quintile thresholds $\qmodel_{\ell,t,g}$ for a given forecast target date $t$ to convert both the ensemble forecast $(\hat{y}_{m,\ell,t,g})_{m=1}^M$ 
and the associated ensemble hindcasts $(\hat{y}_{m,\ell,\delta,t,g})_{m=1}^M$ into input CDF forecasts:
\begin{talign}
F_{\ell,t,g}(k) = \frac{1}{M}\sum_{m=1}^M \indic{\hat{y}_{m,\ell,t,g} < \qmodel_{\ell,t,g}(k)}
\qtext{and}
F_{\ell,t-\delta\,\textup{years},g}(k) = \frac{1}{M}\sum_{m=1}^M \indic{\hat{y}_{m,\ell,\delta,t,g} < \qmodel_{\ell,t,g}(k)}.
\end{talign}

\begin{algorithm}%
  \caption{\perpp}
  \label{alg:perpp}
  \begin{algorithmic}
  	\INPUT target date $\tstar$; quantile bin $k$; 
  	lead time $\lstar$;  
    training set of observed cumulative quantile bin indicators $(\ind_t)_{t\in \trainset}$ and cumulative probability   forecasts $(\fcst_{\lstar,t})_{t\in \trainset\cup\{\tstar\}}$
    \\
    \INIT forecast period length $L = 7$, \# climatology years $Y=20$ \\
    \STATE // Form rolling probabilistic climatology from observed indicators
    \FOR{target and training dates $t \in \trainset \cup \{\tstar\}$}
    \STATE
    $
    \climvec_{t}(k)
    = \textup{mean}(\{\ind_{s}(k) : \monthday(s) = \monthday(t), 1 \leq \myyear(t)-\myyear(s) \leq Y\})
    $ \\[2pt]
    \ENDFOR
    \STATE // Combine probabilistic forecast, climatology, and lagged observations
    \FOR{grid points $g = 1$ {\bfseries to} $G$}
    \STATE
    $
    \mbi{\hat{\beta}}_g \in \argmin_{\mbi{\beta}} 
    \sum_{t \in \trainset}
    (O_{t,g}(k) - {\mbi{\beta}}^\top{[1,C_{t,g}(k), O_{t-\lstar-L+1,g}(k), O_{t-2\lstar-L+2, g}(k), F_{\lstar,t,g}(k)]})^2
    $ \\[2pt]
    \ENDFOR
    \OUTPUT $\clip\Big(
        \left(\mbi{\hat{\beta}}_g^\top
        [1, C_{\tstar,g}(k), O_{\tstar-\lstar-L+1,g}(k), O_{\tstar-2\lstar-L+2,g}(k), F_{\lstar,\tstar,g}(k)]\right)_{g=1}^G,
        [0, 1]
    \Big)$
  \end{algorithmic}
\end{algorithm}

\subsubsection{Projection and ensembling}

It it possible for the quantile bin predictions of \debpp and \perpp to violate the constraint that cumulative probabilities must be non-decreasing across quantile thresholds. 
In such cases, we use isotonic regression \citepmethods{best1990active} to project their output forecasts onto the space of valid CDFs.  Remarkably, this operation is guaranteed to improve forecast skill as, by minimizing squared error to the set of valid CDFs, it increases RPSS for all possible observations simultaneously. 
The final PBC forecast averages the two projected corrections to capture the complementary benefits of each.

\subsubsection{MicroDuet}

To form the MicroDuet model %
used in the AI Weather Quest, we considered three combinations of the PBC-ECMWF and PBC-PoET forecasts---an equal-weighted average, full weight on PBC-ECMWF, and full weight on PBC-PoET---and selected the combination that maximized 2024 RPSS for each variable and lead time.
The resulting selection gave equal weight to PBC-ECMWF and PBC-PoET for temperature, full weight to PBC-ECMWF for precipitation, and full weight to PBC-PoET for mean sea level pressure.

\subsection{AI Weather Quest evaluation}
For each target weather variable and model, \cref{fig:aiwq} reports the AI Weather Quest leaderboard scores for the September-October-November contest season~\citep{ecmwf2026aiweatherquestleaderboard}. 
The Quest scored each model against ERA5-T ground truth \citep{hersbach2020era5} using the RPSS 
across all Monday target dates $\testset$ from September 1, 2025 -- November 24, 2025 (inclusive) for week 3 and from September 8, 2025 -- December 1, 2025 (inclusive) for week 4.

\subsection{Forecasting extreme events}
When forecasting extreme highs or extreme lows, each model is used to predict the probability of the top semidecile (measurements verifying above the 95th percentile) or the bottom semidecile (measurements verifying below the 5th percentile) respectively. 
Given a test set $\testset$ of target dates, we measured extreme forecast quality via the Brier skill score (BSS) \citepmethods{brier1950verification},
\begin{align}\label{eq:bss}
    \bss = 1-\frac{\sum_{t\in\testset}\sum_{g=1}^G w_g\,\bscore_{t,g}}{\sum_{t\in\testset}\sum_{g=1}^G w_g\,\bsc_{t,g}} 
    \sstext{for}
    \bscore_{t,g} = (F_{t,g}(k) - O_{t,g}(k))^2 
    \sstext{and}
    \bsc_{t,g} = \left(\frac{k}{K} - O_{t,g}(k)\right)^2,
\end{align}
where $K=20$, $k=1$ for extreme lows, $k=K-1$ for extreme highs, $\bscore_{t,g}$ is termed the Brier score (BS) of the forecast, and $\bsc_{t,g}$ is the BS of a baseline climatological forecast that assigns equal probability to each semidecile bin.

\subsection{Flood forecasting}
To assess forecast skill during observed flood events, we constructed a global catalog of flood occurrences from the Global Disaster Alert and
Coordination System (GDACS; \url{https://www.gdacs.org}).  Flood events
(event type \texttt{FL}) at all alert levels (Green, Orange, Red) were
retrieved through the GDACS REST API for the date range 2016-01-01--2026-05-29.  Each
GDACS event record provides a centroid (latitude, longitude) coordinate and an event start date; from these we derived for every event (i) a
geographic bounding box of $\pm 10\degree$ around the centroid (clamped
to $[-90\degree, 90\degree]$ in latitude) and
(ii) a 7-day target period beginning on the event start date.  
Only
events with start dates matching the target date of an ECMWF dynamical forecast (i.e., Monday or Friday start dates until 2023-07-15 for week 3 or until 2023-07-22 for week 4 and all start dates thereafter) were evaluated. 
This yielded 2435 flood events for week 3 and 2428 for week 4. 
The severe flood analysis filtered these events to floods with Orange or Red alert levels yielding 88 events at each lead time.

For a given test set $\testset$ of flood event target dates, we measured flood forecast quality via the Brier skill score \cref{eq:bss} for the top precipitation semidecile bin ($k=K-1$ for $K=20$) with $g$ now enumerating the bounding box grid cells.
\subsection*{Statistical methods}
Paired Wilcoxon signed-rank tests were used to test for significant RPSS differences across the $T=26$ weeks of the September-October-November and December-January-February AI Weather Quest seasons \citep{talib2026aiweatherquest}. 
Since the exact issuance dates of the FuXi-S2S and ECMWF dynamical model seldom coincided in the 2017--2021 test period of Extended Data \cref{fig:fuxi}, we reported (two-sided) stationary block bootstrap \citepmethods{politis1994stationary} confidence intervals based on 
the bias-corrected and accelerated (BCa) bootstrap \citepmethods{efron2020automatic} with automatic block length selection \citepmethods{patton2009correction}, 
$5000$ bootstrap replicates, $T=514$ target dates for FuXi-S2S, and $T=522$ target dates for PBC-ECMWF.

For significance testing in \cref{fig:extremes3}c, we used paired one-sided bootstrap tests of BSS improvement using the BCa bootstrap with $5000$ bootstrap replicates, $T=88$ severe flood events, $T=2435$ total week $3$ flood events, and $T=2428$ total week $4$ flood events. 
For all other significance tests, we used paired one-sided stationary block bootstrap tests of skill score improvement 
using the BCa bootstrap 
with 
automatic block length selection, 
$5000$ bootstrap replicates, 
and 
$T$ target dates for $T=939$ (Figures~\ref{fig:pbc_ecmwf_bar} and \ref{fig:extremes3}a and Extended Data Figures~\ref{fig:pbc_ecmwf_regional},  \ref{fig:extremes4}a, and \ref{fig:pbc_breakdown}a), 
$T=179$ (\cref{fig:pbc_ai_rpss}a), 
$T=104$ (\cref{fig:pbc_ai_rpss}b and Extended Data Figure~\ref{fig:pbc_breakdown}b, week 3), 
$T=107$ (\cref{fig:pbc_ai_rpss}b and Extended Data Figure~\ref{fig:pbc_breakdown}b, week 4), 
$T=232$ (Extended Data Figure~\ref{fig:pbc_ecmwf_seasonal}, DJF),
$T=237$ (Extended Data Figure~\ref{fig:pbc_ecmwf_seasonal}, MAM),
$T=236$ (Extended Data Figure~\ref{fig:pbc_ecmwf_seasonal}, JJA), and
$T=234$ (Extended Data Figure~\ref{fig:pbc_ecmwf_seasonal}, SON).

\section*{Data availability}
All ERA5 data used in this study is available via the Climate Data Store (CDS) of the Copernicus Climate Change Service (\url{https://cds.climate.copernicus.eu}). 
All dynamical data used in this study is available in the Subseasonal to Seasonal (S2S) Prediction Project Database \citepmethods{vitart2017subseasonal}. 
S2S is a joint initiative of the World Weather Research Programme (WWRP) and the World Climate Research Programme (WCRP). The original S2S database is hosted at ECMWF as an extension of the TIGGE database (\url{https://apps.ecmwf.int/datasets/data/s2s}). 
All AI Weather Quest evaluations are available via the AI Weather Quest leaderboard (\url{https://aiweatherquest.ecmwf.int/leaderboards/}). 
All FuXi-S2S forecasts are available in the official FuXi-S2S forecast dataset \citepmethods{zhongfuxi2025}. 
All flood events are available via  the Global Disaster Alert and
Coordination System REST API (\url{https://www.gdacs.org}).
All maps were generated using the Cartopy Python package\citepmethods{met2010cartopy} with coastlines, continguous United States borders, and lake boundaries provided by the Natural Earth public domain dataset (\url{https://www.naturalearthdata.com/}).
\section*{Code availability}
Open-source code implementing PBC and the experiments in this work is available at \url{https://github.com/mouatadid/pbc} (DOI:10.5281/zenodo.22238796). 
Open-source code for the PoET model is available at
\url{https://github.com/ecmwf-lab/poet/}. Code for training AIFS-SUBS is available in the \texttt{anemoi} open source package at \url{https://github.com/ecmwf/anemoi-core}.

\bibliographystylemethods{sn-mathphys-num} %
\bibliographymethods{refs}%
\backmatter

\subsection*{Funding}
PO was supported by FAPERJ (Fundação Carlos Chagas Filho de Amparo à Pesquisa do Estado do Rio de Janeiro) grant SEI-260003/020673/2025. 
JC was supported by National Science Foundation grant no. AGS-2140909. 
JS and JT were supported with funding from the European Union, provided to ECMWF under the Contribution Agreement between the European Union, represented by the European Commission, and ECMWF on the implementation of the Destination Earth (DestinE) Initiative.

\subsection*{Author contributions}

H.G. performed her work as an intern at Microsoft Research. 
H.G., S.M., P.O., J.C., H.D., Z.N., G.F., A.L., J.A.W., L.M. conceptualized the work. 
S.M., J.C., A.L., J.A.W., L.M. managed the project.
H.G., S.M., P.O., G.F.,  J.T., J.A.W., L.M. curated data. 
H.G., S.M., P.O., H.D., Z.N., J.A.W., L.M. developed models. 
H.G., S.M., P.O., J.B., G.F., J.T., J.S., J.A.W., L.M. wrote code. 
H.G., S.M., P.O., J.B., J.S., J.A.W., L.M. conducted experiments and did evaluation. 
H.G., S.M., P.O., J.C., L.M. wrote the original draft. 
H.D., J.B., G.F., A.L., J.S., J.T., J.A.W., L.M. reviewed and edited the paper.

\subsection*{Competing interests}

H.D., Z.N., A.L., J.A.W., and L.M. are employees of Microsoft. 
All authors declare that they have no competing interests.
\clearpage
\begin{appendices}
\opt{arxiv}{\section{Extended Data}}
\setcounter{figure}{0}
\renewcommand{\figurename}{Extended Data Figure}
\renewcommand{\thefigure}{\arabic{figure}}
\setcounter{table}{0}
\renewcommand{\tablename}{Extended Data Table}
\renewcommand{\thetable}{\arabic{table}}
\opt{arxiv}{\subsection{Baselines}}

\newcommand{\baselinetablecaption}{\textbf{Baselines.} Dynamical, AI, hybrid (dynamical + AI), and post-processing baselines compared with probabilistic bias correction in this work.\label{tab:baselines}}

\opt{arxiv}{
\begin{table}[htbp]
\caption{\baselinetablecaption}
\begin{tabular}{lcc}
\toprule
\textbf{Baseline} & \textbf{Model Type} & \textbf{Evaluation Periods} \\
\midrule
Raw ECMWF \citep{tuppi2023simultaneous} & Dynamical & 2016--2024, 2025 real-time \\
Debiased ECMWF \citep{SUBS-M-climate} & Post-processing, Dynamical & 2016--2024, 2025 real-time \\
FuXi-S2S \citepmethods{zhongfuxi2025} & AI & 2017--2021 \\
PoET \citep{benbouallegue2024improving} & Post-processing, Hybrid & 2024 \\
AIFS-SUBS \citep{schloer2026aifssubs} & AI & 2025, 2025 real-time \\
FengshunHybrid \citepmethods{cmaandfdu2025aiweatherquest} & Hybrid & 2025 real-time \\
LPM \citepmethods{lp2025aiweatherquest} & Post-processing, Hybrid & 2025 real-time \\
Debiased ECCC \citepmethods{vitart2017subseasonal} & Post-processing, Dynamical & 2025 real-time \\
Debiased KMA \citepmethods{vitart2017subseasonal} & Post-processing, Dynamical & 2025 real-time \\
Debiased JMA \citepmethods{vitart2017subseasonal} & Post-processing, Dynamical & 2025 real-time \\
Debiased HMCR \citepmethods{vitart2017subseasonal} & Post-processing, Dynamical & 2025 real-time \\
Debiased NOAA \citepmethods{vitart2017subseasonal} & Post-processing, Dynamical & 2025 real-time \\
Dynamical MME \citepmethods{vitart2017subseasonal} & Post-processing, Dynamical & 2025 real-time \\
\bottomrule
\end{tabular}
\end{table}}

\opt{nature}{
\captionsetup[table]{hypcap=false}
\captionof{table}{\baselinetablecaption}
\nocitemethods{cmaandfdu2025aiweatherquest,lp2025aiweatherquest}
}

\opt{arxiv}{\clearpage\subsection{Seasonal RPSS}}
\begin{figure}[htbp!]
    \centering
    \opt{arxiv}{\includegraphics[width=\linewidth]{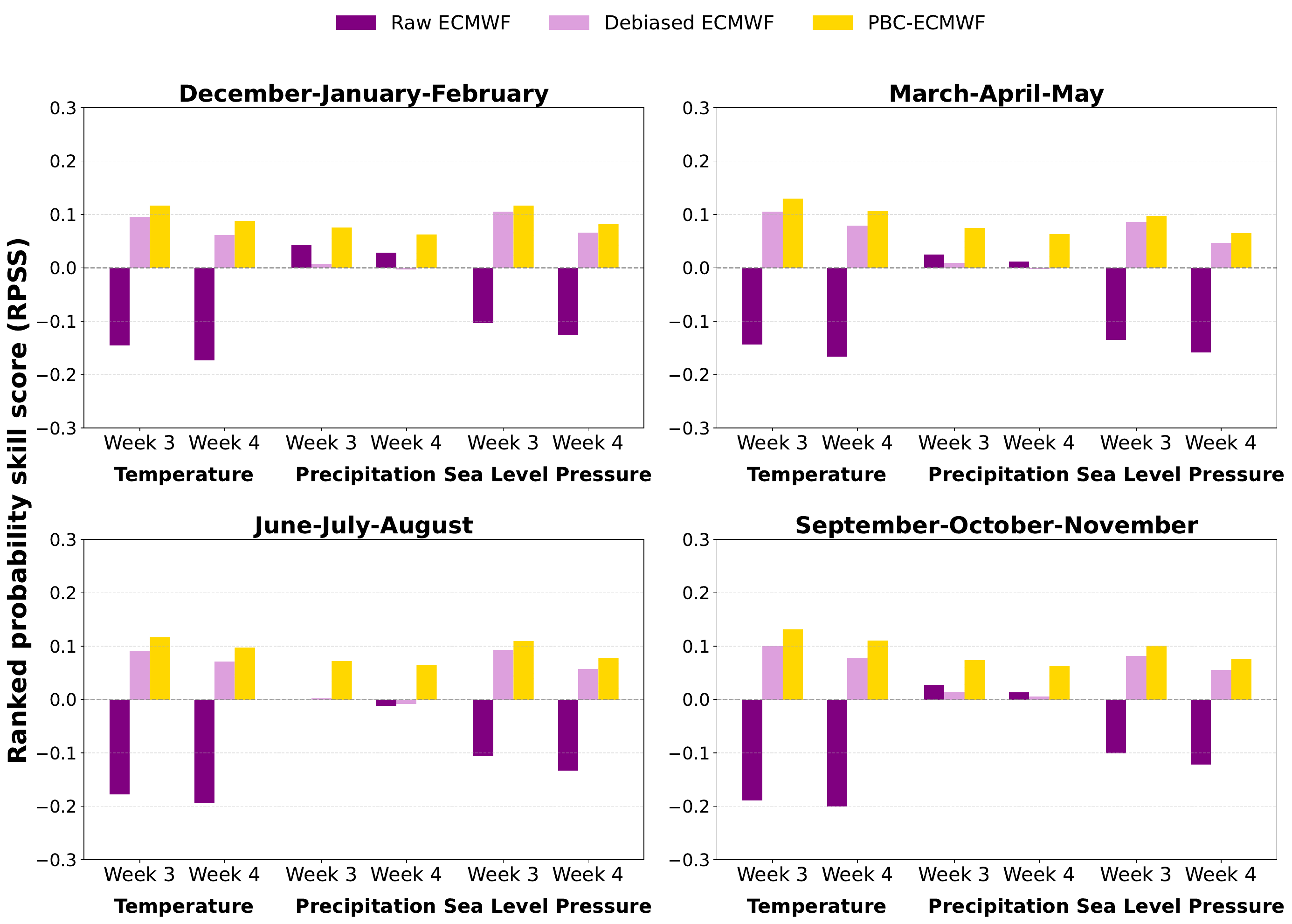}\vspace{\baselineskip}}
    \caption{\textbf{Global forecast skill per season for the leading dynamical model (ECMWF) and its probabilistic bias correction (PBC).} Across 
    the years 2016--2024, PBC boosts the skill of ECMWF and operationally debiased ECMWF in every season. 
    \significance[raw and debiased ECMWF]
    }
    \label{fig:pbc_ecmwf_seasonal}
\end{figure}

\opt{arxiv}{\clearpage\subsection{Regional RPSS}}

\begin{figure}[H]
  \centering
  \opt{arxiv}{\includegraphics[trim=.25cm 0 .5cm 0,clip,width=\textwidth]{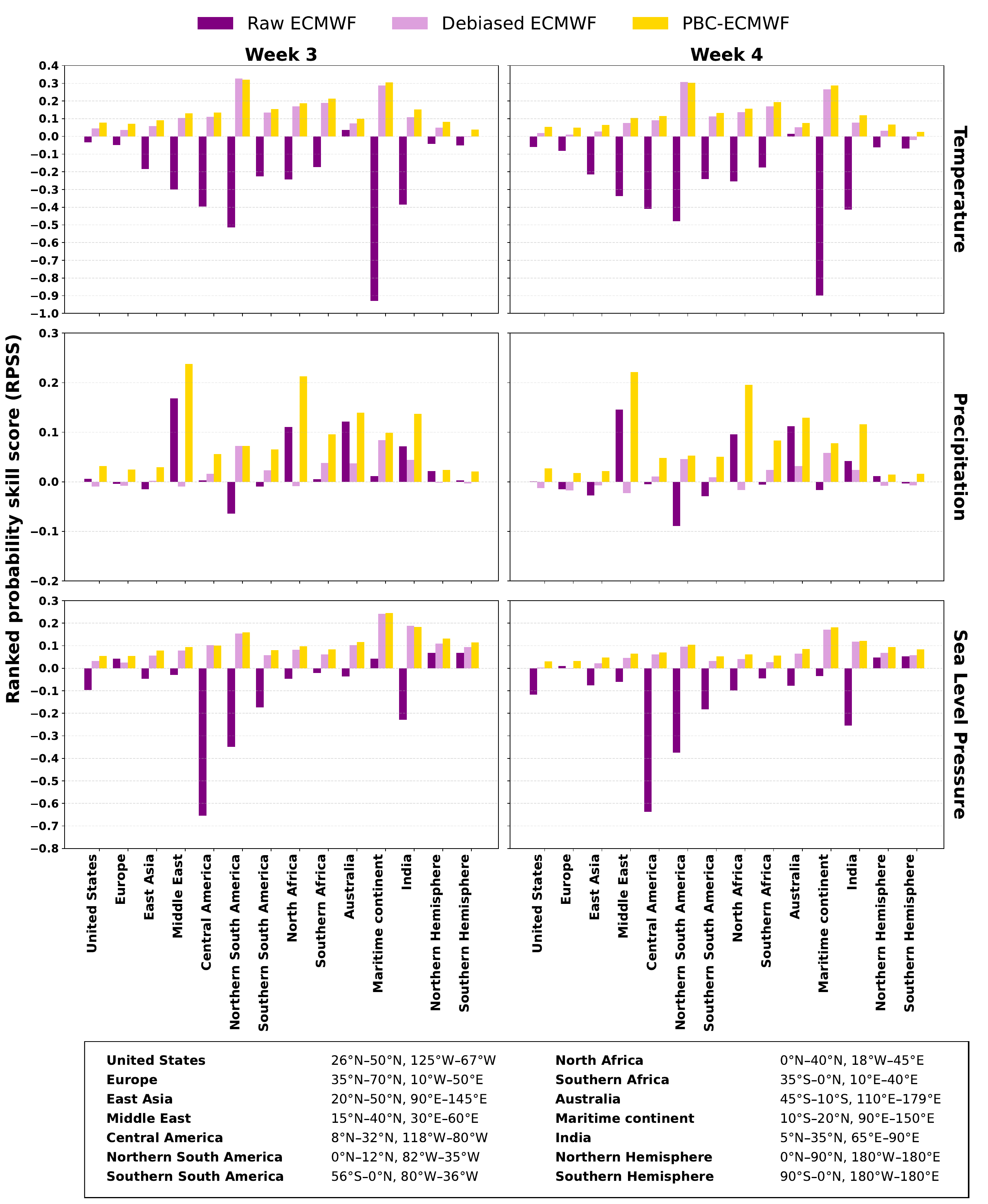}}
  \caption{\textbf{Regional forecast skill of the leading dynamical model (ECMWF) and its probabilistic bias correction (PBC).} Across the years 2016–2024 and world regions defined by the associated latitude-longitude bounding boxes, PBC boosts the skill of ECMWF and operationally debiased ECMWF. \singlesignificance{raw and debiased ECMWF}\vspace{-\baselineskip}
  }
  \label{fig:pbc_ecmwf_regional}
\end{figure}

\opt{arxiv}{\clearpage\subsection{Spatial bias}}

\newcommand{\biascaption}[1]{\textbf{Spatial distribution of #1 model bias for the leading dynamical model (ECMWF)  and its probabilistic bias correction (PBC).} Across the globe, the years 2016--2024, and all probabilistic quintile bins, PBC reduces the systematic bias of ECMWF more effectively than operational debiasing.}

\begin{figure}[H]
  \centering
    \opt{arxiv}{\includegraphics[width=\textwidth]{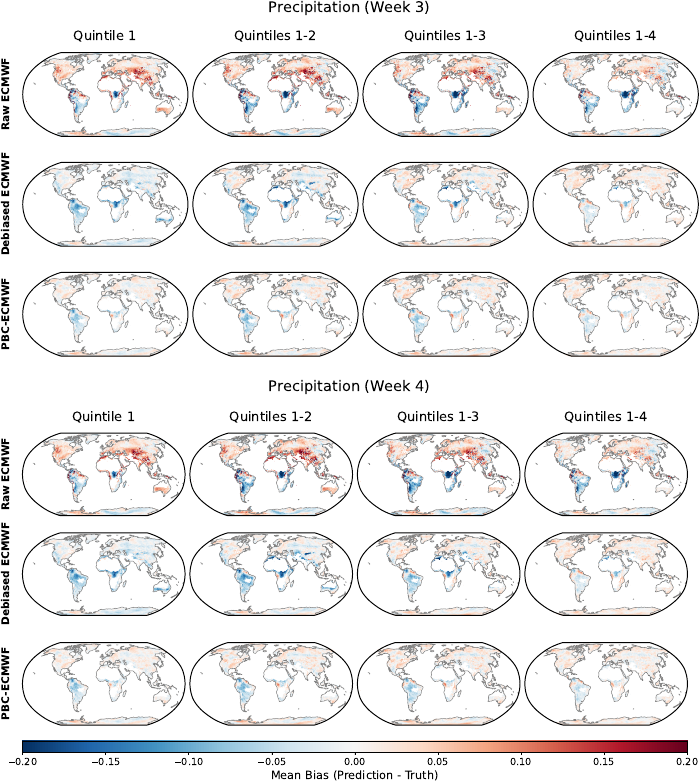}\vspace{\baselineskip}}
  \caption{\biascaption{precipitation} \arid}
  \label{fig:bias_pr}
\end{figure}

\begin{figure}[htpb]
  \centering
    \opt{arxiv}{\includegraphics[width=\textwidth]{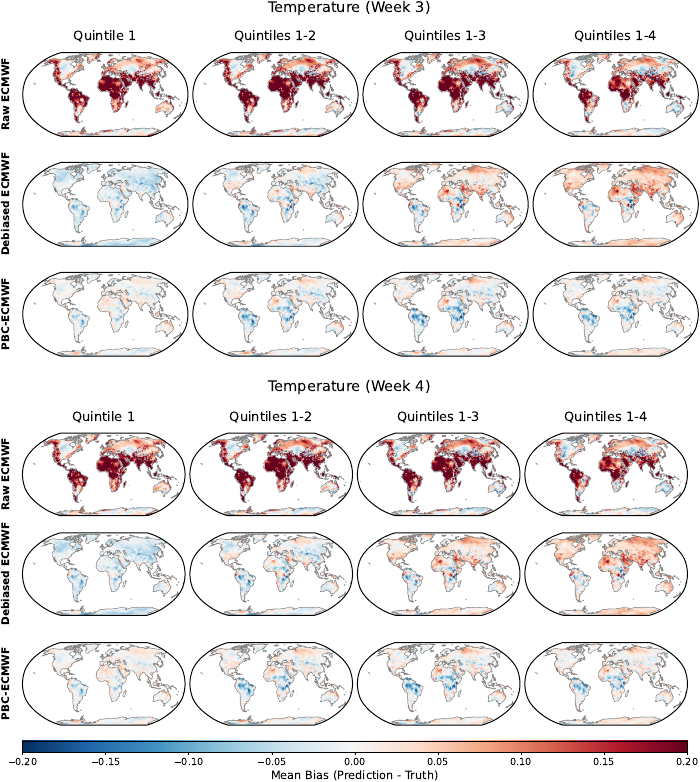}\vspace{\baselineskip}}
  \caption{\biascaption{temperature}}
  \label{fig:bias_tas}
\end{figure}

\begin{figure}[htpb]
  \centering
  \opt{arxiv}{\includegraphics[width=\textwidth]{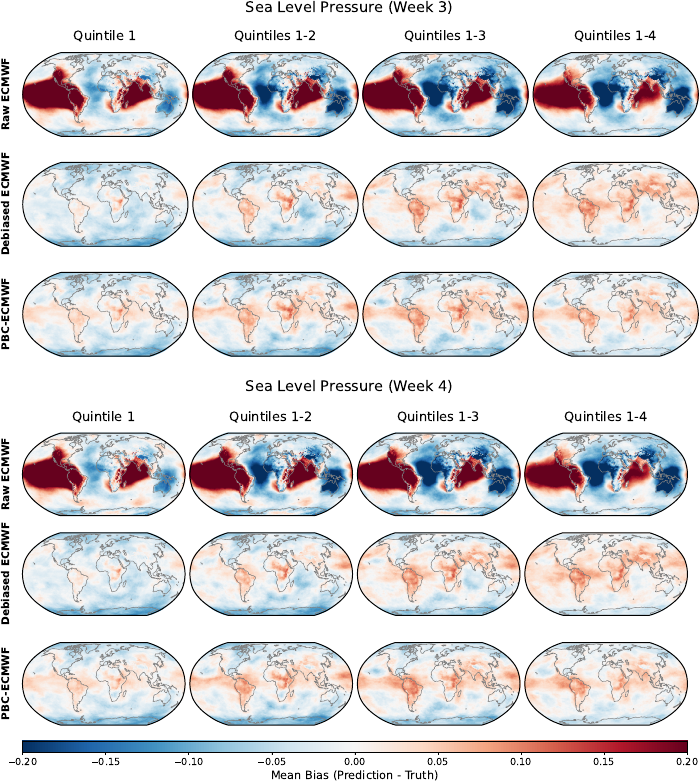}\vspace{\baselineskip}}
  \caption{\textbf{Spatial distribution of mean sea level pressure model bias for the leading dynamical model (ECMWF)  and its probabilistic bias correction (PBC).} Across the globe, the years 2016--2024, and all probabilistic quintile bins, PBC substantially reduces the systematic bias of ECMWF.}
  \label{fig:bias_mslp}
\end{figure}

\opt{arxiv}{\clearpage\subsection{Comparison with FuXi-S2S}}

\begin{figure}[htbp!]
    \centering

    \opt{arxiv}{\includegraphics[width=.7\textwidth]{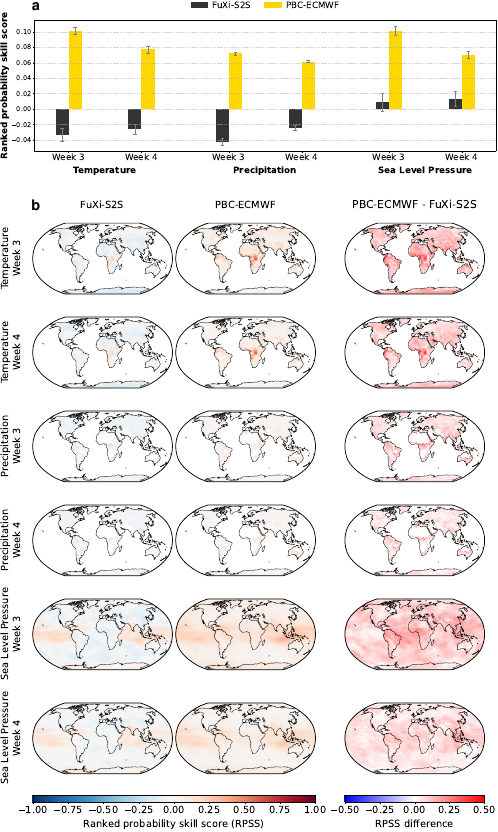}\vspace{\baselineskip}}

    \caption{\textbf{Global forecast skill of FuXi-S2S and PBC-ECMWF.} 
    Across 
    the years 2017--2021, PBC-ECMWF \textbf{(a)} displays a pronounced improvement in skill over the FuXi-S2S AI model for each variable and lead time  
    with \textbf{(b)} 99--100\% of grid cells showing skill gains for temperature, 97--99\% for precipitation, and 96--100\% for mean sea level pressure.
    \arid
    The error bars display 95\% bootstrap confidence intervals.
    }
    \label{fig:fuxi}
\end{figure}

\opt{arxiv}{\clearpage\subsection{Components of PBC-ECMWF and PBC-PoET}}
\begin{figure}[htbp!]
    \centering
    \opt{arxiv}{\includegraphics[width=\textwidth]{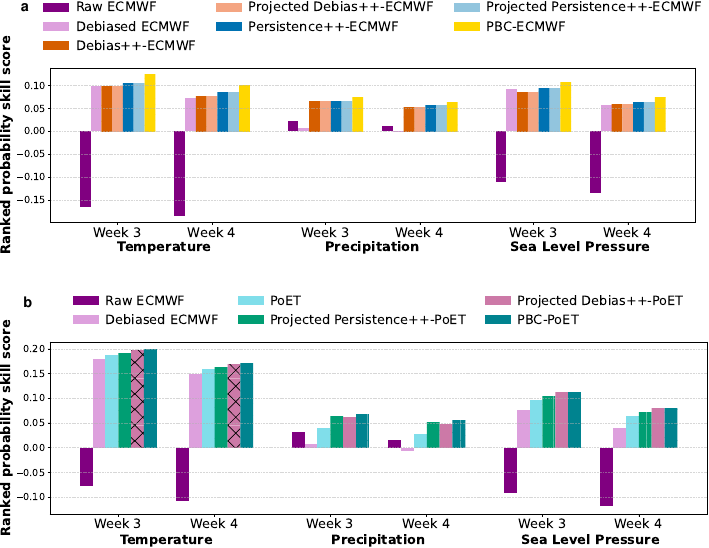}\vspace{\baselineskip}}
    \caption{\textbf{Global forecast skill of the leading dynamical model (ECMWF), the hybrid (AI + dynamical) model PoET, and the components of their probabilistic bias corrections (PBC).} (\textbf{a}) Across the years 2016--2024, for every variable and lead time, \perpp and \debpp individually boost the subseasonal skill of raw ECMWF forecasts and contribute to robust skill gains over operational debiasing through ensembling. 
    (\textbf{b}) In 2024, PoET already improves the skill of its dynamical input, ECMWF. Nevertheless, PBC further enhances skill for every variable and lead time thanks to direct adaptive CDF optimization with \debpp, the integration of recent and climatological observations through \perpp, and the robust combination of these complementary corrections via ensembling. \singlesignificance[at least one baseline]{all baselines}}
    \label{fig:pbc_breakdown}
\end{figure}

\opt{arxiv}{\clearpage\subsection{Extreme events}}

\begin{figure}[htbp]
    \centering
    \opt{arxiv}{\includegraphics[trim=.25cm 0 .25cm 0, clip, width=\linewidth]{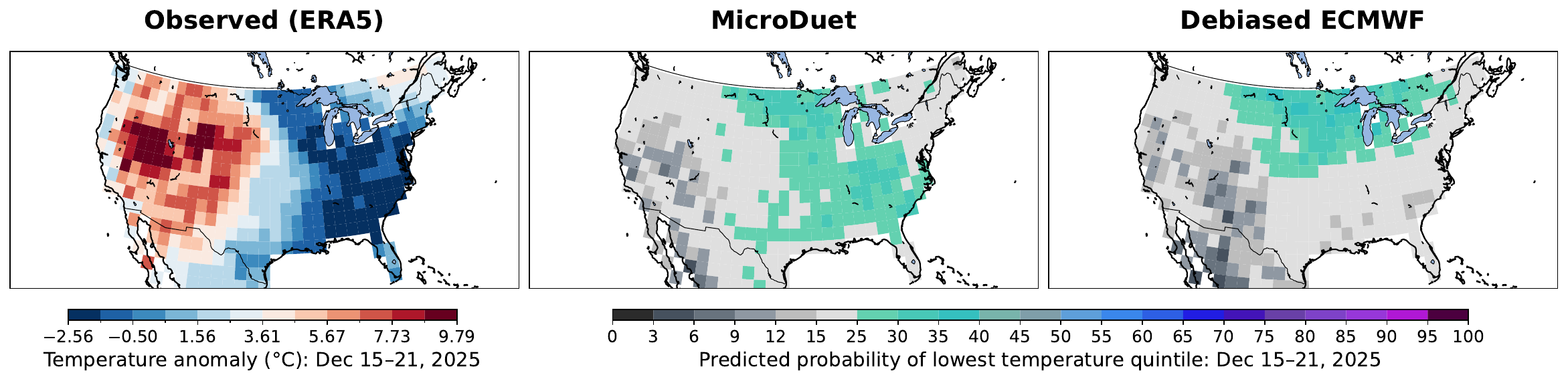}\vspace{\baselineskip}}
    \caption{\textbf{Forecasting the December 2025 cold air outbreak in the Eastern United States.} 
    MicroDuet's real-time week 4 temperature forecast (issued on November 20, 2025 for the AI Weather Quest competition) anticipated the mid-December cold air outbreak in the Eastern United States \citep{oliveira2025northeast}, an extreme cold event \citep{atienza2025polar} missed by the leading debiased dynamical model from ECMWF.} 
    \label{fig:aiwq_extreme}
\end{figure}

\begin{figure}[htbp]
    \centering
    \opt{arxiv}{\includegraphics[width=.8\textwidth]{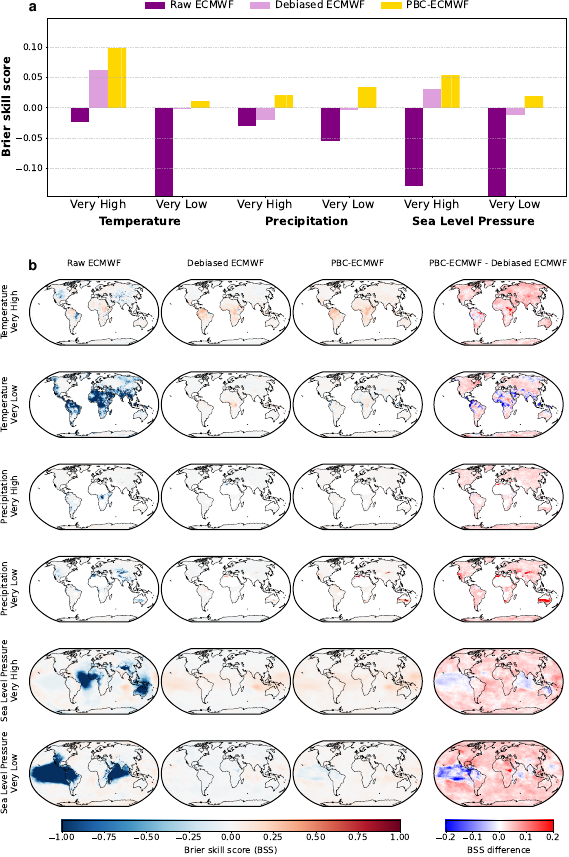}\vspace{\baselineskip}}
    \caption{\textbf{Forecasting extreme weather with the leading dynamical model (ECMWF) and its probabilistic bias correction (PBC) in week 4.} Across the globe and the years 2016--2024, PBC boosts ECMWF extreme forecast skill more effectively than operational debiasing protocols with 96\% of grid cells showing skill gains for precipitation, 90\% for sea level pressure, and 85\% for temperature. 
    (\textbf{a}) \significance[raw and debiased ECMWF] (\textbf{b}) \arid \vspace{-\baselineskip}}
    \label{fig:extremes4}
\end{figure}

\end{appendices}

\end{document}